\documentclass{article} % For LaTeX2e
\usepackage{iclr2022_conference,times}
\usepackage{helvet}
\usepackage{courier}
\usepackage{hyperref}
\usepackage{latexsym}

\usepackage{booktabs}
\usepackage{mathtools}
\usepackage{varwidth}
\usepackage{empheq}
\usepackage{amssymb}
\usepackage{wrapfig}
\usepackage{tabularx}
\usepackage{multirow}
\usepackage{epsfig}
\usepackage{amsmath}

\makeatletter
\g@addto@macro\normalsize{%
  \setlength\abovedisplayskip{1pt}
  \setlength\belowdisplayskip{2pt}
  \setlength\abovedisplayshortskip{1pt}
  \setlength\belowdisplayshortskip{2pt}
}
\makeatother

\usepackage{graphicx}
\usepackage{cancel}
\usepackage{amsfonts}       % blackboard math symbols
\usepackage{nicefrac}       % compact symbols for 1/2, etc.
\usepackage{bbm}
\usepackage{bm}
\usepackage{subcaption}
\usepackage{listings}
\usepackage{enumitem}% http://ctan.org/pkg/enumitem
\usepackage{defs}

\renewcommand{\vec}[1]{\boldsymbol{#1}}
\renewcommand{\vec}[1]{\boldsymbol{#1}}

\usepackage{siunitx} % to round numbers
\renewcommand{\vec}[1]{\boldsymbol{#1}}
\usepackage{algorithm}
\usepackage{algorithmic}
\usepackage{wrapfig}
\usepackage{microtype}

\usepackage{amsmath,amsfonts,bm}

\def\eqref#1{equation~\ref{#1}}
\def\1{\bm{1}}

\DeclareMathAlphabet{\mathsfit}{\encodingdefault}{\sfdefault}{m}{sl}
\SetMathAlphabet{\mathsfit}{bold}{\encodingdefault}{\sfdefault}{bx}{n}

\usepackage{hyperref}
\usepackage{url}

\title{Deep Neural Compression Via Concurrent Pruning and Self-Distillation}

\author{James O' Neill \\
Department of Computer Science, \\
University of Liverpool \\
Liverpool L69 3BX, UK \\
\texttt{james.o-neill@liverpool.ac.uk} \\
\AND
Sourav Dutta \& Haytham Assem \\
Huawei Ireland Research Center \\
Georges Court, Townsend St \\
Dublin 2, D02 R156, Ireland \\
\texttt{\{sourav.dutta2,haytham.assem\}@huawei.com} 
}

\iclrfinalcopy % Uncomment for camera-ready version, but NOT for submission.
\begin{document}

\maketitle

\begin{abstract}
%Pruning reduces the parameter count in neural networks and aims to maintain close to the performance of the unpruned network. The pruning criteria can be task-agnostic (e.g weight magnitude) or task-dependent (e.g gradient-based pruning). 

%In this work, we argue that enforcing a pruned network to be similar to its originally fine-tuned (and unpruned) network has generalization benefits and recovers faster from performance degradation directly after iterative pruning steps. 

% Although, gradient-based pruning methods reveal which weights are important for a given task, they do not constrain the resulting sparse network to have similar representations to the original unpruned network. For high compression rates, this can lead to pruning instability. 

% SOURAV- Added first line
Pruning aims to reduce the number of parameters while maintaining performance close to the original network. 
This work proposes a novel \emph{self-distillation} based pruning strategy, whereby the representational similarity between the pruned and unpruned versions of the same network is maximized. Unlike previous approaches that treat distillation and pruning separately, we use distillation to inform the pruning criteria, without requiring a separate student network as in knowledge distillation.
We show that the proposed {\em cross-correlation objective for self-distilled pruning} implicitly encourages sparse solutions, naturally complementing magnitude-based pruning criteria. 
Experiments on the GLUE %~\citep{wang2018glue} 
and XGLUE %~\citep{liang2020xglue} 
benchmarks show that self-distilled pruning increases mono- and cross-lingual language model performance.
% in the high compression regime and sets a new state of the art for iterative pruning methods.
Self-distilled pruned models also outperform smaller Transformers with an equal number of parameters and are competitive against (6 times) larger distilled networks.
%as the knowledge distilled baselines. 
We also observe that self-distillation (1) maximizes class separability, (2) increases the signal-to-noise ratio, and (3) converges faster after pruning steps, providing further insights into why self-distilled pruning improves generalization. 
% improves convergence and can be further improved by using our proposed cross-correlation distilled pruning objective. 
\end{abstract}

\section{Introduction}
\vspace*{-1mm}
% In recent years, deep neural networks (DNNs) have shown impressive performance~\cite{schmidhuber2015deep,vaswani2017attention,conneau2019unsupervised}. However, improved performance has come at the cost of higher storage requirements, memory footprint, computing resources and energy utilization~\cite{cheng2017survey,deng2020model}. 

% ,han2015compressing,li2016pruning
Neural network pruning~\citep{mozer1989skeletonization,karnin1990simple,reed1993pruning} zeros out weights of a pretrained model with the aim of reducing parameter count and storage requirements, while maintaining performance close to the original model. The criteria to choose which weights to prune has been an active research area over the past three decades~\citep{karnin1990simple,lecun1990optimal,han2015compressing,anwar2017structured,molchanov2017variational}. 
Lately, there has been a focus on pruning models in the transfer learning setting whereby a self-supervised pretrained model trained on a large amount of unlabelled data is fine-tuned to a downstream task while weights are simultaneously pruned. \iffalse This setup is referred to as \emph{fine-pruning}.\fi In this context, recent work proposes to learn important scores over weights with a continuous mask and prune away those that having the smallest scores~\citep{mallya2018piggyback,sanh2020movement}.
% Prior work use important scores to learn which weights to prune for a given task throughout iterative pruning~\cite{mallya2018piggyback}.

However, these learned masks double the number of parameters in the network, requiring twice the number of gradient updates to tune the original parameters \emph{and} their continuous masks~\citep{sanh2020movement}. Ideally, we aim to perform task-dependent fine-pruning \emph{without} adding more parameters to the network, or at least far lesser than twice the count. More generally, we desire pruning methods that can recover from performance degradation directly after pruning steps, faster than current pruning methods while encoding task-dependent information into the pruning process. % This recovery becomes more salient as the network becomes highly compressed.
To this end, we hypothesize self-distillation may recover performance faster after consecutive pruning steps, which becomes more important with larger performance degradation at higher compression regime. Additionally, self-distillation has shown to encourage sparsity as the training error tends to 0~\citep{mobahi2020self}. This implicit sparse regularization effect complements magnitude-based pruning criteria.

Hence, this paper proposes to {\em maximize the cross-correlation} between output representations of the fine-tuned pretrained network and a pruned version of the same network -- referred to as \emph{self-distilled pruning} (SDP). Unlike typical knowledge distillation (KD) where the student is a separate network trained from random initialization, here the student is initially a masked version of the teacher.
% as the KD objective forces the pruned network to have similar representations to the unpruned network
We focus on pruning fine-tuned monolingual \emph{and} cross-lingual transformer models, namely BERT~\citep{devlin2018bert} and XLM-RoBERTa~\citep{conneau2019unsupervised}. To our knowledge, this is the first study that introduces the concept of {\em self-distilled pruning} and analyze iterative pruning in cross-lingual contexts.
%of cross-lingual language models. Below we summarize 
In a nutshell, our main contributions are as follows:
%
% We formalize the connection between the pruning criteria and regularization induced by self-distillation in our work.  

% Unlike, prior work that treats pruning and knowledge distillation as independent steps in a compression pipeline, we simultaneously use self-distillation to improve iterative pruning while having the computation benefit of not requiring an additional student network.
%Moreover, iterative pruning can lead to catastrophic forgetting of the original network which makes it difficult for the network to recover from directly after each pruning step. In other terms, after a number of iterative pruning steps, the pruning instability grows larger and therefore requires either a larger interval between successive pruning steps or a mechanism which minimizes instability directly after a pruning step, particularly as the model is heavily pruned. Thus, a method that improves convergence of pruned networks would improve the overall as it reduces the amount of training required to recover from growing instability.
%This is desirable as KD has shown to improve convergence, which when used with iterative pruning, translates to recovering from the aforementioned instability directly after a pruning step. 

%  All prior work studies of compression methods in natural language processing have only focused on monolingually trained language models~\citep{wang2019structured,sanh2020movement,liu2018efficient,gordon2020compressing}

%\vspace{-.75em}
%\paragraph{Contributions}%\mbox{}\\
\begin{enumerate}[leftmargin=*]
\itemsep0em
    % \textbf{1)}
    \item We propose \emph{self-distilled pruning}, a novel pruning framework that improves the generalization of pruned networks \emph{without} introducing any additional parameters, using only a set of soft targets.
    % \textbf{2)} 
    \item Inspired by the recent success of correlation-based objectives for representation learning in computer vision~\citep{zbontar2021barlow}, we propose the use of a {\em cross-correlation objective for self-distillation pruning} that reduces redundancy and encourages sparse solutions, naturally fitting with magnitude-based pruning. This sets state of the art results for magnitude-based pruning.  
    % works by maximizing cross-correlation on the diagonal of the cross-correlation matrix and reduces redundancy by minimizing the off-diagonals of the cross-correlation matrix. This objective encourages a sparse solution which naturally fits well with magnitude-based pruning criteria. We establish a new state of the art for unstructured pruning, while outperforming competing self-distillation losses.  
    % \textbf{3)}
    \item We provide three insights as to why self-distillation leads to more generalizable pruned networks. Namely, we observe that self-distilled pruning (1) {\em recovers performance faster} after pruning steps (i.e., improves convergence), (2) {\em maximizes the signal-to-noise ratio} (SNR), where pruned weights are considered as noise, and (3) {\em improves the fidelity} between pruned and unpruned representations as measured by mutual information of the respective penultimate layers. 
    \item A comprehensive study of iterative pruning for monolingual and cross-lingual pretrained models on GLUE and XGLUE benchmarks. To our knowledge, this is the only work to include an evaluation of pruned model performance in the cross-lingual transfer setting.
\end{enumerate}
%This opens up a new line of research towards simultaneously combining knowledge distillation and pruning in a single retraining pass.
% Namely, we show that self-distillation improves optimization by smoothening the loss surface due to more informative posteriors given by a fine-tuned network, as opposed to one-hot targets.  
% \item We find that self-distilled pruned models score higher for a chosen complexity measure which is shown to correlate well generalization.  
% \item Although we find that self-distilled pruning generally improves test performance, it leads to poorer calibration of output probabilities when compared to pruning without a self-distilled objective. 
% ~\citeauthor{gotmare2018closer} have empirically observed that the dark knowledge transferred by the teacher is localized mainly in higher layers and does not affect early (feature extraction) layers much.
% ~\citeauthor{furlanello2018born} interprets dark knowledge as importance weighting.~\citeauthor{dong2019distillation} shows that early-stopping is crucial for reaching dark-knowledge of self-distillation.~\citeauthor{abnar2020transferring} empirically study how inductive biases are transferred through distillation. Ideas similar to self-distillation have been used in areas besides modern machine learning but with different names such as diffusion and boosting in both the statistics and image processing communities~\cite{milanfar2012tour}.
% In the proceeding sections we shed light on potential reasons why self-distilled models generalize better and how these insights pertain to improving pruned models. 
\vspace{-1em}
\section{Background and Related Work}
%\paragraph{Pruning}
%Pruning weights is used to reduce the number of parameters in a pretrained DNN with the aim of reducing storage, model runtime and maintaining the same, or close to the performance original unpruned network. Iterative pruning prunes while retraining until the desired network size and accuracy tradeoff is met. Pruning is not carried out at random, but selected so that unimportant information about past experiences is discarded. 

\textbf{Regularization-based pruning} can be achieved by using a weight regularizer that encourages network sparsity. Three well-established regularizers are $L_1$, $L_2$ and $L_0$ weight regularization~\citep{louizos2017learning,liu2017learning,ye2018rethinking} for weight sparsity ~\citep{han2015learning,han2015compressing}. For structured pruning, Group-wise Brain Damage~\citep{lebedev2016fast} and SSL~\citep{wen2016learning} propose to use Group LASSO~\citep{yuan2006model} to prune whole structures (e.g., convolution blocks or blocks within standard linear layers).% where the penalty strength is still kept in small scale because the penalty is uniformly applied to all the weights.
%To resolve this,~\citeauthor{ding2018auto} and~\cite{wang2019eigendamage} propose to employ different penalty factors for different weights, enabling large regularization.
~\citet{park2020lookahead} aim to avoid pruning small weights if they are connected to larger weights in consecutive layers and vice-versa, by constraining the Frobenius norm of the pruned layers to be close to unpruned network. 

\textbf{Importance-based pruning} assigns a score for each weight in the network and removes weights with the lowest importance score. The simplest scoring criteria is magnitude-based pruning (MBP), which uses the lowest absolute value (LAV) as the criteria~\citep{reed1993pruning,han2015learning,han2015compressing} or $L_1/L_2$-norm for structured pruning~\citep{liu2017learning}. MBP can be seen as a zero-th order pruning criteria. However higher order pruning methods approximate the difference in pruned and unpruned model loss using a Taylor series expansion up until $1^{st}$ order ~\citep{lecun1990optimal,hassibi1993second} or the 2$^{nd}$ order, which requires approximating the Hessian matrix~\citep{martens2015optimizing,wang2019eigendamage,singh2020woodfisher} for scalability. Lastly, the regularization-based pruning is commonly used with importance-based pruning e.g using $L_2$ weight regularization alongside MBP.

% \textbf{Other model compression methods.} Apart from pruning, there are also many other model compression approaches, e.g., quantization~\cite{courbariaux2016binarized,rastegari2016xnor}, knowledge distillation ~\cite{bucilua2006model,hinton2015distilling}, low rank decomposition ~\cite{denton2014exploiting,jaderberg2014speeding,lebedev2014speeding,zhang2015efficient} and efficient architecture design or search~\cite{howard2017mobilenets,sandler2018mobilenetv2,howard2019searching,he2016deep,zhang2015efficient,tan1905rethinking,zoph2016neural,elsken2019neural}. They are orthogonal to network pruning and can work with the proposed methods to compress more.

\textbf{Knowledge Distillation} (KD) transfers the logits of an already trained network~\citep{hinton2015distilling} and uses them as soft targets to optimize a student network. The student network is typically smaller than the teacher network and benefits from the additional information soft targets provide. There has been various extensions that involve distilling intermediate representations~\citep{romero2014fitnets}, distributions~\citep{huang2017like}, maximizing mutual information between student and teacher representations~\citep{ahn2019variational}, using pairwise interactions for improved KD~\citep{park2019relational} and contrastive representation distillation~\citep{tian2019contrastive,neill2021semantically}.

\textbf{Self-Distillation} is a special case of KD whereby the student and teacher networks have the same capacity. Interestingly, self-distilled students often generalize better than the teacher~\citep{furlanello2018born,yang2019training}, however the mechanisms by which self-distillation leads to improved generalization remains somewhat unclear. Recent works have provided insightful observations of this phenomena. For example,~\cite{stanton2021does} have shown that soft targets make optimization easier for the student when compared to the task-provided one-hot targets.~\cite{allen2020towards} view self-distillation as implicitly combining ensemble learning and KD to explain the improvement in test accuracy when dealing with multi-view data. The core idea is that the self-distillation objective results in the network learning a unique set of features that are distinct from the original model, similar to features learned by combining the outputs of independent models in an ensemble. Given this background on pruning and distillation, we now describe our proposed methodology {\em SDP}.
% ~\citeauthor{phuong2019towards} have already established that convergence towards the expected risk is faster when using knowledge distillation (KD), where a student network learns from the output logits~\cite{hinton2015distilling} or intermediate representations~\cite{romero2014fitnets} of a pretrained network. We suspect that this is because the teacher network soft labels provide more informative targets than the typical one-hot targets by implicitly informing the class boundaries, leading to better class separation. 
%

\section{Proposed Methodology}
% In this section, we describe our proposed SDP, a cross-correlation self-distillation objective and a theoretical analysis of how SDP generalizes by increasing signal-to-noise ratio and mutual information between pruned and unpruned versions of the network. 

We begin by defining a dataset $\mathcal{D}: = \{(X_i, y_i)\}^D_{i=1}$ with single samples $s_i = (X_i, \vec{y}_i)$, where each $X_i$ (in the $D$ training samples) consists of a sequence of vectors $X_i := (\vec{x}_1, \ldots, \vec{x}_N)$ and $\vec{x}_i \in \mathbb{R}^{d}$. For structured prediction (e.g., NER, POS) $y_i \in \{0, 1\}^{N \times C}$, and for single and pairwise sentence classification, $y_i \in \{0, 1\}^{C}$, where $C$ is the number of classes.
Let $\vec{y}^S=f_{\theta}(X_i)$ be the output prediction ($y^{S} \in \mathbb{R}^{C}$) from the student $f_{\theta}(\cdot)$ with pretrained parameters $\theta:= \{\mat{W}_l, \vec{b}_l\}_{l=1}^{L}$ for $L$ layers. The input to each subsequent layer is denoted as $\vec{z}_{l} \in \mathbb{R}^{n_{l}}$ where $\vec{x} \coloneqq \vec{z}_0$ for $n_l$ number of units in layer $l$ and the corresponding output activation as $\vec{A}_{l} = g(\vec{z}_{l})$. 
The loss function for standard classification fine-tuning is defined as the cross $\ell_{CE}(\vec{y}^S, \vec{y}):= -\frac{1}{C}\sum_{i=1}^c \vec{y}_c\log(\vec{y}^s_c)$.% where for a single sample $s_i$, $\mathcal{L}: \mathcal{Y} \times \mathbb{R}^n \to \mathbb{R}$. 

For self-distilled pruning, we also require an already fine-tuned teacher network $f_{\Theta}$, that has been tuned from the pretrained state $f_{\theta}$, to retrieve the soft teacher labels $y^{T} := f_{\Theta}(\vec{x})$, where $y^{T} \in \mathbb{R}^C$ and $\sum_c^C y^{T}_c=1$. The soft label $\vec{y}^{T}$ can be more informative than the one-hot targets $\vec{y}$ used for standard classification as they implicitly approximate pairwise class similarities through logit probabilities.
% After fine-tuning $f_{\theta}$ through stochastic gradient descent (SGD), the output activations of $f_{\Theta}$ are used to perform self-distillation.
The Kullbeck-Leibler divergence $\ell_{\mathrm{KLD}}$ is then used with the main task cross-entropy loss $\ell_{CE}$ to express $\ell_{\mathrm{SDP-KLD}}$ as shown in \autoref{eq:kld_loss}, 
\vspace*{1mm}
\begin{equation}\label{eq:kld_loss}
\ell_{\mathrm{SDP-KLD}} = (1 - \alpha) \ell_{\mathrm{CE}}(\vec{y}^{S}, \vec{y}) + \alpha \tau^2 D_{\mathrm{KLD}}\big(\vec{y}^{S}, \vec{y}^{T}\big) 
\end{equation}

where $D_{\mathrm{KLD}}(\vec{y}^{S},\vec{y}^{T}) = \mathbb{H}( \vec{y}^T) - \vec{y}^{T} \log(\vec{y}^{S})$, $\mathbb{H}(\vec{y}^T)=\vec{y}^{T}\log(\vec{y}^{T})$ is the entropy of the teacher distribution and $\tau$ is the softmax temperature. Following~\cite{hinton2015distilling}, the weighted sum of cross-entropy loss and KLD loss shown in \autoref{eq:kld_loss} is used as our main SDP-based KD loss baseline, where $\alpha \in [0, 1]$. 
After each pruning step during iterative pruning, we aim to recover the immediate performance degradation by minimizing $\ell_{\mathrm{SDP-KLD}}$. In our experiments, we use weight magnitude-based pruning as the criteria for SDP given MBP's flexibility, scalability and miniscule computation overhead (only requires a binary tensor multiplication to be applied for each linear layer at each pruning step). However, $D_{\mathrm{KLD}}$ only distils the knowledge from the soft targets which may not propagate enough information about the intermediate dynamics of the teacher, nor does it penalize representational redundancy. This brings us to our proposed cross-correlation SDP objective.

% As we will discuss in the proceeding sections (\textbf{say which subsection}), gradient magnitudes are less effective when using self-distillation as the self-distilled objective activates gradients more equally which may not task-specific. 

\iffalse
\begin{equation}
\frac{\partial W_i(\tau)}{\partial \tau} = - \frac{\partial L}{\partial W_i}
(W_1(\tau), \ldots ,W_N (\tau))
\end{equation}
The student is trained until convergence, i.e. $\tau \to \infty$. We measure the transfer risk of the trained student, defined as the probability that its prediction differs from that of the teacher, $R(h) = P_{x \sim P}x[h(x) \neq h_{*}(x)]$.
\fi

\vspace{-.5em}
\subsection{Maximizing Cross-Correlation Between Pruned and Unpruned Embeddings}
\vspace{-.5em}
Iterative pruning can be viewed as progressively adding noise $\mat{M}_l \in \{0, 1\}^{n_{l-1}\times n_l}$ to the weights $\mat{W}_l \in \mathbb{R}^{n_{l-1}\times n_l}$. 
Thus, as the pruning steps increase, the outputs become noisier and the relationship between the inputs and outputs becomes weaker. Hence, a correlation measure is a natural choice for dealing with such pruning-induced noise. To this end, we use a cross-correlation loss to maximize the correlation between the output representations of the last hidden state of the pruned network and the unpruned network to reduce the effects of this pruning noise. The proposed {\em cross-correlation SDP loss function}, $\ell_{\mathrm{CC}}$, is expressed in \autoref{eq:cc}, where $\lambda$ controls the importance of minimizing the non-adjacent pairwise correlations between $z^S$ and $z^T$ in the correlation matrix $\mathcal{C}$. Here, $m$ denotes the sample index in a mini-batch of $M$ samples. Unlike $\ell_{\mathrm{KLD}}$, this loss is applied to the outputs of the last hidden layer as opposed to the classification logit outputs. Thus, we have, 
\vspace*{1mm}
\begin{equation}\label{eq:cc}
\ell_{\mathrm{CC}} := \sum_i (1 - \mathcal{C}_{ii})^2 + \lambda \sum_i \sum_{j\neq i} \mathcal{C}_{ij}^2 \quad s.t, \quad
\mathcal{C}_{ij} := \frac{\sum_{m} \vec{z}_{m,i}^{S} \vec{z}_{m,j}^{T} }{\sqrt{\sum_m (\vec{z}_{m,i}^{S})^2}\sqrt{\sum_m(\vec{z}_{m,j}^{T})^2 }}
\end{equation}

Maximizing correlation along the diagonal of $\mathcal{C}$ makes the representations {\em invariant to pruning noise}, while minimizing the off-diagonal term {\em decorrelates the components} of the representations that are batch normalized.  
% Another common point between the two losses is that they both rely on batch statistics to measure this variability. However, the INFONCE objective maximizes the variability of the embeddings by maximizing the pairwise distance between all pairs of samples, whereas our method does so by decorrelating the components of the embeddings vectors. The embeddings are normalized along the batch dimension using batch normalization.
To reiterate, $\vec{z}^S$ is obtained from the pruned version of the network ($f_{\Theta^p}$) and $\vec{z}^{T}$ is obtained from the unpruned version ($f_{\Theta}$). 
\begin{wrapfigure}[15]{r}{0.5\textwidth}
\vspace{-1em}
    \centering
    \includegraphics[scale=0.55]{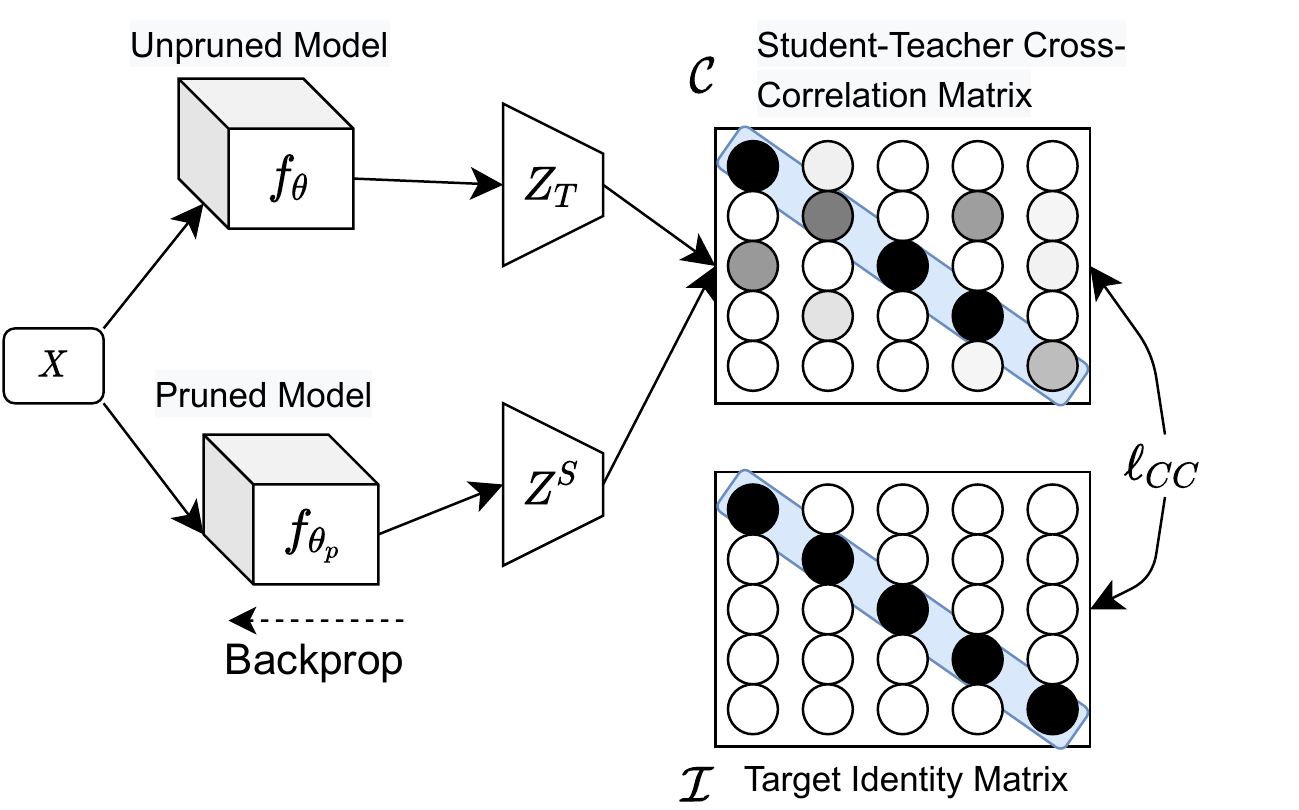}
    \vspace{-1.25em}
    \caption{\textbf{Self-Distilled Pruning w/ a Cross-Correlation Knowledge Distillation Loss.}}
    \label{fig:sdp_cc}
\end{wrapfigure}

Since the learned output representations should be similar if their inputs are similar, we aim to address the problem where a correlation measure may produce representations that are instead \emph{proportional} to their inputs. To address this, we use batch normalization across mini-batches to stabilize the optimization when using the cross-correlation loss, essentially avoiding local optima that correspond to degenerate representations that do not distinguish proportionality. In our experiments, this is used with the classification loss and KLD distillation loss 
%$\ell_{\mathrm{KLD}}$ %SOURAV - Equation ref changed from 1 to 3
as shown in \autoref{eq:sdpcc}.
\vspace*{1mm}
\begin{equation}\label{eq:sdpcc}
\ell_{\mathrm{SDP-CC}} = (1 - \alpha) \ell_{\mathrm{CE}}(\vec{y}^{S}, \vec{y}) + \alpha^2  \ell_{\mathrm{KLD}}(\vec{y}^{S}, \vec{y}^T) + \beta \ell_{\mathrm{CC}}(\vec{z}^{S}, \vec{z}^{T})
\end{equation}
To bring it all together, in \autoref{fig:sdp_cc} we show the proposed framework of {\em Self-Distilled Pruning with cross-correlation loss} (SDP-CC), where $\mathcal{I}$ is the identity matrix. Additionally, we provide a PyTorch based pseudo-code for SDP-CC in \autoref{alg:sdp} for a single epoch, in the general case. We see that the inner training loop requires little additional overhead, only requiring the extra computation (compared to normal pruning) to compute \texttt{distil\_loss} and \texttt{cross\_corr\_loss}.  
% Here $\mat{A}$ and $\mat{A}_p$ are the activation outputs of the last layer and $\alpha$ controls the influence of this SDP loss relative to the $\ell_{\mathrm{CE}}$ loss. 
\begin{wrapfigure}[25]{R}{0.5\textwidth}
\begin{minipage}{0.48\textwidth}
\vspace{-3em}
\begin{algorithm}[H]
   \caption{PyTorch pseudo-code of SDP-CC.}
   \label{alg:sdp}
   
    \definecolor{codeblue}{rgb}{0.25,0.5,0.5}
    \lstset{
      basicstyle=\fontsize{7.2pt}{7.2pt}\ttfamily\bfseries,
      commentstyle=\fontsize{7.2pt}{7.2pt}\color{codeblue},
      keywordstyle=\fontsize{7.2pt}{7.2pt},
    }
\begin{lstlisting}[language=python]
# model: student transformer 
# teacher_model: fine-tuned transformer
# alpha, beta: distillation loss weights
# method: chosen pruning criteria 
# prune_rate: compression amount in [0, 1] 
# lambda: weight on the off-diagonal terms
for x in loader: # batch loader of N samples
    # input_ids,attention_id & label_ids.
    x = tuple(xs.to(device) for xs in x)
    inputs = {"input_ids": x[0],
    "attention_mask": x[1], "labels": x[2]}
    outputs = student_model(**inputs)
    teacher_outs = teacher_model(**inputs)
    distil_loss = F.kld_divergence_loss(
        x[2], teacher_outs[2])
    # inputs last hidden representation for
    # [CLS] token and applies Equation 2. 
    cross_corr_loss = cc_loss(outputs[1][-1],
    teacher_outs[1][-1], lambda)
    # loss given as first element in tuple    
    loss = outputs[0] + alpha * distil_loss 
    + beta * cross_corr_loss  
    # gradient clipping
    torch.nn.utils.clip_grad_norm_(
    model.parameters(), max_grad_norm)
    # compute backprop and update gradients
    loss.backward()
    optimizer.step()
# apply pruning after a whole epoch
with torch.no_grad():
    prune_method(model, method, prune_rate)
\end{lstlisting}
\end{algorithm}
\end{minipage}
\end{wrapfigure}

\subsection{A Frobenius Distortion Perspective of Self-Distilled Pruning}
To formalize the objective being minimized when using MBP with self-distillation, we take the view of {\em Frobenius distortion minimization}~\citep[FDM;][]{dong2017learning} which says that layer-wise MBP is equivalent to minimizing the Frobenius distortions of a single layer. 
This can be described as $\min_{\mat{M}:||\mat{M}||_0=p} ||\mat{W} - \mat{M} \odot \mat{W} ||_F$, where $\odot$ is the Hadamard product and $p$ is a constraint of the number of weights to remove as a percentage of the total number of weights for a layer. Therefore, the output distortion is approximately the product of single layer Frobenius distortions.
However, this minimization only defines a 1$^{st}$ order approximation of pruning induced Frobenius distortions which is a loose approximation for deep networks. In contrast, the $\vec{y}^T$ targets provide higher-order information outside of the $l$-th layer being pruned in this FDM framework because $\Theta$ encodes information of all neighboring layers. Hence, we reformulate the FDM problem for SDP as an approximately higher-order MBP method as in \autoref{eq:fdm_sdp} where $\mat{W}^{T}$ are the weights in $f_{\Theta}$.

\begin{equation}\label{eq:fdm_sdp}
\min_{\mat{M}:||\mat{M}||_0=p} \Big[||\mat{W} - \mat{M} \odot \mat{W} ||_F + \lambda||\mat{W}^{T} - \mat{M} \odot \mat{W} ||_F \Big]    
\end{equation}

As described in~\cite{dong2017learning,hassibi1993second}, the error function can be approximated with a Taylor Series (TS) expansion as $\mathcal{E}(\mat{Z}_l^T) - \mathcal{E}(\mat{Z}_l) = \delta \mathcal{E}_l \approx \big(\frac{\partial \mathcal{E}_l}{\partial\mathbf{\theta}^l}\big)^{\top}  \delta \mathbf{\theta}_l + \frac{1}{2}\delta \mathbf{\theta}_l^T \mat{H}_l \delta \mathbf{\theta}_l + O(
||\delta\mathbf{\theta}_l||^3)$. Since, MBP corresponds to the $1^{st}$ order term of this TS expansion, we can express the TS approximation of the error function $\mathcal{E}(\cdot)$ with self-distilled pruning as shown in \autoref{eq:error_diff}, where $\mat{Z}^*_l$ is the true latent representation of the input. 
\vspace*{1mm}
\begin{equation}\label{eq:error_diff}
    2\mathcal{E}(\mat{Z}_l) - \mathcal{E}(\mat{Z}^*_l) - \mathcal{E}(\mat{Z}_l^T) = \delta \mathcal{E}_l \approx \Big(\frac{\partial \mathcal{E}_l}{\partial\mathbf{\theta}_l}\Big)^{\top}  \delta \mathbf{\theta}_l + \lambda \Big(\frac{\partial \mathcal{E}_l^T}{\partial\mathbf{\theta}_l}\Big)^{\top}  \delta \mathbf{\theta}_l 
\end{equation}

\subsection{How Does Self-Distillation Improve Pruned Model Generalization ?}\label{sec:sdp_improves_perf}
%Below, we describe three factors that explain improved pruned model generalization when using SDP.
%self-distilled pruning.soft targets used in self-distilled pruning may lead to better generalization in pruned networks. 
% Concretely, we track representational similarity between the penultimate layers of pruned and unpruned networks over consecutive pruning epochs and also show that self-distilled representations maximizes inter-class distances and minimize intra-class distances. This implicit class separation induced through self-distillation leads to faster convergence which allows SDP networks to recover faster from performance degradation directly after pruning steps when compared to pruned network that do not use a self-distillation objective.
% the success of self-distillation in iterative pruning. In short, self-distillation leads to faster convergence \iffalse(\autoref{sec:convergence})\fi, implicit maximization of the signal-to-noise ratio \iffalse(\autoref{sec:snr})\fi and its the performance-fidelity tradeoff between the pruned model and teacher \iffalse(\autoref{sec:fidelity})\fi. 
We put forth the following insights as to the advantages provided by self-distillation for better model generalization, and later experimentally demonstrate their validity.

\textbf{Recovering Faster From Performance Degradation After Pruning Steps.}\label{sec:convergence}
The first explanation for why self-distillation leads to better generalization in iterative pruning is that the soft targets bias the optimization and smoothen the loss surface through implicit similarities between the classes encoded in the logits. We posit this too holds true for performance recovery after pruning steps, as the classification boundaries become distorted due to the removal of weights. Faster convergence is particularly important for high compression rates where the performance drops become larger. 

% We posit that the reason why iterative pruning benefits from our proposed self-distillation is that the model converges faster directly after a pruning step by when given more informative logits or intermediate representations from the teacher. In this subsection we formalize the mathematical framework which describes the improvement in pruned model convergence and its relationship to SDP.

\vspace{.25em}
\textbf{Implicit Maximization of the Signal-to-Noise Ratio.}\label{sec:snr}
One explanation for faster convergence is that optimizing for soft targets translates to maximizing the margin of class boundaries given the implicit class similarities provided by teacher logits. Intuitively, task provided one-hot targets do not inform SGD of how similar incorrect predictions are to the correct class, whereas the teacher logits do, to the extent they have learned on the same task.
To measure this, we use a formulation of the signal-to-noise ratio\footnote{A measure typically used in signal processing to evaluate signal quality.} (SNR) to measure the class separability and compactness differences between pruned model representations trained with and without self-distillation. 
We formulate SNR as \autoref{eq:snr}, where for a batch of inputs $\mat{X}$, we obtain $\mat{Z}$ output representations from the pruned network, which contain samples with $C$ classes where each class has the same $N$ number of samples. The numerator expresses the average $\ell_2$ inter-class distance between instances of each class pair and the denominator expresses the intra-class distance between instances within the same class.
% 1/C cancels on top and bottom
\vspace{2mm}
\begin{equation}\label{eq:snr}
    \mathrm{SNR}(\mat{Z})= \frac{1/N(C-1)^2 \sum_{n}^{N}\sum_{c=1}^{C}\sum_{i\neq c}^{C}||\sqrt{\mat{Z}_{c, n}} - \sqrt{\mat{Z}_{i, n}}||_2}{1/C (N-1)^2\sum_{c=1}^C\sum_{n}^{N}\sum_{j \neq n}||\sqrt{\mat{Z}_{c, n}} - \sqrt{\mat{Z}_{c,j}}||_2} 
\end{equation}

This estimation is $C-1\binom{C+1}{2}$ in the number of pairwise distances to be computed between the inter-class distances for the classes. For large output spaces (e.g., language modeling) we recommend defining the top $k$-NN classes for each class and estimate their distances on samples from them.
%top-k classes.
% as instances of distant classes will not be informative about the class boundaries.
%However, we did not have to do so for our set of experiments and use all pairwise inter-class distances.

\vspace{0.25em}
\textbf{Quantifying Fidelity Between Pruned Models Trained With and Without Self-Distillation.}\label{sec:fidelity} A natural question to ask is \textit{how much generalization power does the distilled soft targets provide when compared to the task provided one-hot targets ?} If best generalization is achieved when $\alpha=1$ in \autoref{eq:kld_loss}, this implies that the pruned network should have as high fidelity as possible with the unpruned network. However, as we will see there is a bias-variance trade-off between fidelity and generalization performance, i.e., $\alpha=1$ is not optimal in most cases. To measure fidelity between SDP representations and standard fine-tuned representations, we compute their {\em mutual information} (MI) and compare this to the MI between representations of pruned models without self-distillation and standard fine-tuned models.
% We denote the representations of the penultimate layer of $f^T$ over the whole batch of samples as $Z^T = (\vec{z}^S_1, \vec{z}^S_2 \ldots, \vec{z}^S_n)$ and we denote the representations of $f^{T}_p$ that is pre-initialized with $f^{T}$ as $Z^S = (\vec{z}^T_1, \vec{z}^T_2 \ldots, \vec{z}^T_n)$.
The MI between continuous variables can be expressed as,
\vspace*{1mm}
\begin{equation}\label{eq:mi}
    \hat{I}(\mat{Z}^T ; \mat{Z}^S) = \mathrm{H}(\mat{Z}^T) - \mathrm{H}(\mat{Z}^T|\mat{Z}^S) = - \mathbb{E}_{\mat{z}^T}[\log p(\mat{Z}^T)] + \mathbb{E}_{\mat{Z}^T,\mat{Z}^S}[\log p(\mat{Z}^T|\mat{Z}^S)]
\end{equation}
%\begin{equation}
%I(Z^T; Z^S) = \mathrm{H}(Z^S) - \mathrm{H}(Z^S|Z^T),    
%\end{equation}
where $\mathrm{H}(\mat{Z}^T)$ is the the entropy of the teacher representation and $\mathrm{H}(\mat{Z}^{T}|\mat{Z}^{S})$ is the conditional entropy that is derived from the joint distribution $p(\mat{Z}^{T}, \mat{Z}^{S})$. 
%where $\mathrm{H}(Z^S)$ is the entropy of student representations $Z^S$ and $\mathrm{H}(Z^T|Z^S)$ is the conditional entropy. 
This can also be expressed as the KL divergence between the joint probabilities and product of marginals as $I(Z^T; Z^S) = D_{\mathrm{KLD}}[p(Z^S, Z^T)|| p(Z^{S})p(Z^{T})]$. However, these theoretical quantities have to be estimated from test sample representations. We use a $k$-NN based MI estimator~\citep{kraskov2004estimating,evans2008computationally,ver2013information,ver2000non} which partitions the supports into a finite number of bins of equal size, forming a histogram that can be used to estimate $\hat{I}(Z^S; Z^T)$ based on discrete counts in each bin. This MI estimator is given as,
% Empirically, the joint distribution $p(\mat{Z}^{T}, \mat{Z}^{S})$ is a result of aggregation over the layers with input $X$ sampled from the input distribution.
\begin{equation}
    I(z^S; z^T) \approx \epsilon \Big(\log \frac{\phi_{[\vec{z}^S]}(i, k_{[\vec{z}^S]})\phi_{[\vec{z}^T]}(i, k_{[\vec{z}^T]})}{\phi_z(i, k)}\Big)
\end{equation}
where $\phi_{z^{S}}(i, k_{[z^{S}]})$ is the probability measure of the $k$-th nearest neighbour ball of $\vec{z}^{S} \in \mathbb{R}^{n_L}$ and $\omega_{[\vec{z}^T]}(i, k_{[\vec{z}^T]})$ is the probability measure of the $k_y$-th nearest neighbour ball of $\vec{z}^{T} \in \mathbb{R}^{n_L}$ where $n_L$ is the dimension of the penultimate layer. In our experiments, we use 256 bins for the histogram with Gaussian smoothing and $k=5$ (see ~\cite{kraskov2004estimating} for further details).
% Assume some metrics to be given on the spaces spanned by $X,Y$ and $Z=s_X,Y_d$. We can then rank, for each point $z_i= (x_i , y_i)$, its neighbors by distance $d_{i,j}=||\vec{z}_{i} - \vec{z}_j||: d_{i,j_1} \leq d_{i,j_2} \leq d_{i,j_3} \leq \ldots$. Similar rankings can be done in the subspaces $X$ and $Y$. The basic idea of [20–22] is to estimate $H(X)$ from the average distance to the k-nearest neighbor, averaged over all $x_i$. Mutual information could be obtained by estimating in this way $H(X)$, $H(Y)$, and $H(X,Y)$ separately and using [1] $I(X,Y) = H(X) + H(Y) - H(X,Y)$.  But this would mean that the errors made in the individual estimates would presumably not cancel, and therefore we proceed differently. Indeed we will present two slightly different algorithms, both based on the above idea. Both use for the space $Z=(X,Y)$ the maximum norm, $||z - z'|| = \max \{||x - x'||,||y - y'||\}$, while any norms can be used for $||x - x'||$ and $||y - y'||$. Let us denote by $\epsilon_x(i)/2$ the distance from $z_i$ to its $k$th neighbor, and by $\epsilon_x(i)/2$ and $\epsilon_y(i)/2$ the distances between the same points projected into the $X$ and $Y$ subspaces. Obviously, $\epsilon(i)=\max\{\epsilon_x(i), \epsilon_y(i)\}$.

%\vspace{-1em}
\section{Experimental Setup}
\vspace{-1em}
\textbf{Datasets.}
We perform experiments on monolingual tasks within the GLUE~\citep{wang2018glue} benchmark\footnote{WNLI is excluded for known issues, see the Q. 12 on the \hyperlink{https://gluebenchmark.com/faq}{GLUE benchmark FAQ}.} with pretrained BERT$_{\mathrm{Base}}$ and multilingual tasks from the XGLUE benchmark~\citep{liang2020xglue} with pretrained XLMR$_{\mathrm{Base}}$. In total, this covers 18 different datasets, covering pairwise classification, \iffalse(QAM, QADSM, WPR, RTE, )\fi sentence classification\iffalse(NC,)\fi, structured prediction and question answering. To our knowledge, this work is the first to analyse iterative pruning in the context of cross-lingual models and their application on multilingual datasets. %Further dataset statistics can be found in supplementary material. 

\textbf{Iterative Pruning Baselines. }%\mbox{}\\
For XGLUE tasks, we perform 15 pruning steps on XLM-RoBERTA$_{\mathrm{Base}}$, one per 15 epochs, while for the GLUE tasks, we perform 32 pruning steps on BERT$_{\mathrm{Base}}$. The compression rate and number of pruning steps is higher for GLUE tasks compared to XGLUE, because GLUE tasks involve evaluation in the {\em supervised classification} setting; whereas in XGLUE we report in the more challenging {\em zero-shot cross-lingual transfer} setting with only a single language used for training (i.e., English). At each pruning step, we uniformly pruning 10\% of the parameters for both the models. Although prior work suggests non-uniform pruning schedules (e.g., cubic schedule~\citep{zhu2017prune}), we did not see any major differences to uniform pruning.% in preliminary experiments. 
We compare the performance of the proposed SDP-CC method against the following baselines:
\begin{itemize}[leftmargin=*]
\itemsep-.05em 
\item \textbf{Random Pruning} ({\em MBP-Random}) - prunes weights uniformly at random across all layers. Random pruning can be considered as a lower bound on iterative pruning performance.
\item \textbf{Layer-wise Magnitude Based Pruning} ({\em MBP}) - for each layer, prunes weights with LAV.
\item \textbf{Global Magnitude Pruning} ({\em Global-MBP}) - prunes LAV weights anywhere in the network.
%\item \textbf{Global Gradient Magnitude Pruning} (Gradient-MBP) - prunes the weights with the lowest absolute value of (weight x gradient), evaluated on a batch of inputs.
\item \textbf{Layer-wise Gradient Magnitude Pruning} ({\em Gradient-MBP}) - for each layer, prunes the weights with LAV of the accumulated gradients evaluated on a batch of inputs.
\item \textbf{$1^{st}$ Taylor Series Pruning} ({\em TS}) - prunes weights based on the LAV of $|$gradient $\times$ weight$|$. 
%onsider the loss function $\ell(w)$ at any time during the training process, we can write a small perturbation at $\vec{w}$ as $\delta \ell = \ell(\vec{w} + \delta \vec{w}) - \ell(\vec{w}) \approx \nabla_w \ell\delta \vec{w} + \frac{1}{2} \delta \vec{w}^{\top} \mat{H} \delta \vec{w}$ where $\nabla_w \ell\delta \vec{w}$ and $\frac{1}{2} \delta \vec{w}^{\top} \mat{H} \delta \vec{w}$ the first and $2^{nd}$ Taylor expansion of $\delta \ell$.
%\item \textbf{LayerDrop} - Drops whole structured modules in the Transformer and prunes those which have the least effect in pruned model performance. 
\item $L_0$ \textbf{norm MBP}~\citep{louizos2017learning} - uses non-negative stochastic gates that choose which weights are set to zero as a smooth approximation to the non-differentiable $L_0$-norm. 
\item $L_1$ \textbf{norm MBP}~\citep{li2016pruning} - applies $L_1$ weight regularization and uses MBP.
\item \textbf{Lookahead pruning (LAP)}~\citep{park2020lookahead} - prunes weight paths that have the smallest magnitude across blocks of layers, unlike MBP that does not consider neighboring layers. 
\item \textbf{Layer-Adaptive MBP (LAMP)}~\citep{lee2020layer} - adaptively compute the pruning ratio per layer. 
\end{itemize}

\vspace*{-2mm}
For all above pruning methods we exclude weight pruning of the embeddings, layer normalization parameters and the last classification layer, as they play an important role for generalization and account for less than 1\% of weights in both BERT and XLM-R$_{\mathrm{Base}}$.

% For a given task, we fine-tune the pre-trained model for the same number of updates (between 6 and 10 epochs) across pruning methods. We follow~\cite{zhu2017prune} and use a cubic sparsity scheduling for Magnitude Pruning (MaP), Movement Pruning (MvP), and Soft Movement Pruning (SMvP). 

% Adding a few steps of cool-down at the end of pruning empirically improves the performance especially in high sparsity regimes.

% We compare our results against several state-of-the-art pruning baselines: Reweighted Proximal Pruning (RPP)~\cite{guo2019reweighted} combines re-weighted $L_1$ minimization and Proximal Projection~\cite{parikh2014proximal} to perform unstructured pruning. LayerDrop~\cite{fan2019reducing} leverages structured dropout to prune models at test time. For RPP and LayerDrop, we report results from authors.
\vspace*{-3mm}
\begin{itemize}[leftmargin=*]
\itemsep-.05em 
\item \textbf{Knowledge Distillation} --
We also compare against a class of smaller knowledge distilled versions of BERT model with varying parameter sizes on the GLUE benchmark. We report prior results of {\em DistilBERT}~\citep{sanh2019distilbert} and also mini-BERT models including {\em TinyBERT}~\citep{jiao2019tinybert}, {\em BERT-small}~\citep{turc2019well} and {\em BERT-medium}~\citep{turc2019well}. 

% MobileBERT~\cite{sun2020mobilebert}
\item \textbf{Self-Distilled Pruning Variant} -- In addition, we consider maximizing the cosine similarity between pruned and unpruned representations in the SDP loss, as $\ell_{\mathrm{SDP-COS}} := \alpha\ell_{\mathrm{CE}}(\vec{y}^{S}, \vec{y}) + \beta \big(1 - \frac{\mat{z}^S \cdot \vec{z}^T}{||\vec{z}^S|| ||\vec{z}^T||}\big)$.
%\autoref{eq:cos_kd}. \begin{equation}\label{eq:cos_kd}
% \end{equation}
 Unlike cross-correlation, there is no decorrelation of non-adjacent features in both representations for SDP-COS. This helps identify whether the redundancy reduction in cross-correlation is beneficial compared to the correlation loss that does not directly optimize this. 
% We also include a Pearson Correlation loss as another scale invariant loss, $\ell_{\mathrm{SDP-PC}} := \alpha\ell_{\mathrm{CE}}(\vec{y}^{S}, \vec{y}) + \beta \big(1 - \frac{\sum_i^m (\mat{Z}_i^S - \bar{\vec{z}}^{S})(\mat{Z}_i^T - \bar{\vec{z}}^{T})}{\sqrt{\sum_i^m (\mat{Z}_i^S - \bar{\vec{z}}^{S})^2}\sqrt{\sum_i^m (\mat{Z}_i^T - \bar{\vec{z}}^{T})^2}}\big)$. 
\end{itemize}
\textbf{Hyperparameter Settings.}
For {\em SDP-KLD}, we tested $\alpha=[0.1, 0.2, 0.5, 0.8, 1]$ on three tasks for GLUE and XGLUE and extrapolate the best performing setting for the remaining tasks of both benchmarks with a fixed $\tau=0.9$. We find $\alpha=0.5$ to perform the best in all cases. For {\em SDP-CC}, we perform tests with $\beta=[10^{-6}, 2\times10^{-5}, 5\times 10^{-5}, 10^{-4}]$, finding that $\beta=2\times10^{-5}$ results in the best average performance. For SDP-COS, we set $\beta=0.05$.  

% Comparison on the development sets of the GLUE benchmark. ELMo, BERT and DistilBERT results are reported by authors.
%ELMo & 68.7 & 44.1  & 91.5 & 68.6 & 76.6 & 70.4 & 86.2 & 53.4 & 71.1 & 56.3 \\
%BERT-base & 79.5 & 56.3 & 92.7 & 86.7 & 88.6 & 89.0 & 89.6 & 69.3 & 91.8 & 53.5 \\
%DistilBERT & 77.0 & 51.3 & 86.9 & 82.2 & 87.5 & 91.3 & 88.5 & 59.9 & 89.2 & 56.3 \\

%\subsection{Quantifying Pruning Performance}
%In the literature there is no principled framework to evaluate iterative pruning methods, criteria etc. Given that the most important region of the test performance versus \% of weights pruned curve is at the inflection point when performance begins to degrade, we propose the following measure. 

\vspace{-.5em}
\section{Empirical Results}
\vspace{-.5em}
\paragraph{Pruning Results on {\em GLUE}.}~\autoref{tab:glue_prune} shows the test performance across all GLUE tasks of the different models with varying pruning ratios, up to {\em 10\% remaining weights} of original BERT$_{\mathrm{Base}}$ along with mini-BERT models~\citep{sanh2019distilbert,turc2019well} of varying size. % that are smaller than BERT$_{\mathrm{Base}}$~\cite{sanh2019distilbert,turc2019well}.
However, for the CoLA dataset, we report at 20\% pruning as nearly all compression methods have an MCC score of 0, making the compressed method performance indistinguishable. For this reason, the GLUE score (\textbf{Score}) is computed for all tasks and methods @10\% apart from CoLA. The best performing compression method per task is marked in {\bf bold}.
We find that our proposed SDP approaches (all three variants) outperform against baseline pruning methods, with {\em SDP-CC} performing the best across all tasks. We note that for the tasks with fewer training samples (e.g., CoLA has 8.5k samples, STS-B has 7k samples and RTE has 3k samples), the performance gap is larger compared to BERT$_{\mathrm{Base}}$, as the pruning step interval is shorter and less training data allows lesser time for the model to recover from pruning losses and also less data for teacher model to distil in the case of using SDP. 
\begin{table}[t]
    \begin{center}
        % \scriptsize
        \resizebox{.95\linewidth}{!}{
        \begin{tabular}[b]{lc|cccccccc}
        \toprule[1.25pt]
        \textbf{Compression Method} & \textbf{Score} & \multicolumn{2}{c}{\underline{\textbf{Single Sentence}}} &  \multicolumn{3}{c}{\textbf{\underline{Similarity and Paraphrase}}} & \multicolumn{3}{c}{\underline{\textbf{Natural Language Inference}}} \\
        
        & (avg.) & \textbf{CoLA} &  \textbf{SST-2} & \textbf{MNLI} & \textbf{MRPC} & \textbf{STS-B} & \textbf{QQP} & \textbf{RTE} & \textbf{QNLI} \\
        
        &  & (mcc) & (acc) & (acc)  &  (f1/acc) & (pears./spear.) & (f1/acc) & (acc) & (acc) \\
        \midrule
        BERT$_\text{Base}$ (Ours) & 84.06 & 53.24 & 90.71 & 80.27 & 80.9/77.7 & 83.5/83.8 & 83.9/88.0 & 68.59 & 86.91 \\ 
         \midrule
         \multicolumn{10}{l}{\textbf{Knowledge Distilled Baselines} (\% parameters w.r.t. original BERT)} \\
         \midrule
        %  82.32
        % 77.0
        DistilBERT (60\%) & 82.85 &  51.3 & 91.3 & 82.2 & 87.5/-.- & 86.9/-.- & -.-/85.5 & 59.9 & 89.2  \\ 
        % 73.5
        BERT-Medium	(44.4\%) & 81.54 & 38.0 & 89.6 & 80.0\iffalse/79.1\fi & 86.6/81.6 & 80.4/78.4 & 69.6/87.9 & 62.2 & 87.7  \\
        % 71.2
        BERT-Small (20\%) & 79.02\iffalse79.92\fi & 27.8 & 89.7 & 77.6\iffalse/77.0\fi & 83.4/76.2 & 78.8/77.0 & 68.1/87.0 & 61.8 & 86.4 \\
        % Mobile-BERT (13.8\%) & 77.0 & 44.1 & 92.6 & 82.5/81.8 & 86.4/- & 80.4/- & 71.3/- & 66.6 & 87.7  \\
        % MNLI-(m/mm) QQP  QNLI SST-2 CoLA  STS-B  MRPC RTE    Avg
        % 82.5/81.8   71.3 87.7 92.6  44.1  80.4   86.4 66.6   77.0

        % 65.8
        BERT-Mini (10\%) & 76.97 & 0.0 & 85.9 & 75.1\iffalse/71.8\fi & 74.8/74.3 &  75.4/73.3 & 66.4/86.2 & 57.9 & 84.1  \\
        %& & CoLA & SST-2 & MRPC &  MNLI-m/MNLI-mm  & STS-B & QQP & RTE & QNLI & WNLI \\
        
        % 64.2 
        BERT-Tiny (3.6\%) & 73.32 & 0.0 & 83.2 & 70.2\iffalse/ 70.3\fi & 81.1/71.1 & 74.3/73.6 & 62.2/83.4 & 57.2 & 81.5 \\ % & 62.3

        \midrule
        \midrule
        %\multirow{1}{*}{\textbf{Pruning Baselines}} & & \\
        \textbf{Pruning Baselines} & & 20\% & 10\% & 10\% & 10\% & 10\% & 10\% & 10\% & 10\% \\

        % & 79.5 & 56.3 & 92.7 & 86.7 & 88.6 & 89.0 & 89.6 & 69.3 & 91.8 & 53.5 \\
        \midrule
    % Acc/F1 for MRPC
    Random & 66.03 & 6.50 & 78.44 & 69.55 & 77.5/67.1 &  27.4/26.9 & 77.07/81.86 & 52.70 & 74.66  \\
    $L_0$-MBP  & 77.25 & 31.68 & 83.37 & 75.61 & 78.4/68.2 & 75.9/75.7  & 81.56/86.49 & \textbf{{64.26}} & 82.62  \\
    $L_2$-MBP  & 76.48 & 29.51 & 83.37 & 76.19 & 78.4/68.2 &  75.3/75.6 & 77.50/82.98 & 62.09 & 82.61 \\
    $L_2$-Global-MBP  & 77.16 & 29.25 & 82.83 & 76.40 & 81.2/69.9 &  75.1/75.5 & 82.77/86.70 & 62.01 & 82.24 \\
    $L_2$-Gradient-MBP & 74.84 & 15.46 & 82.91 & 72.51 & 81.0/73.7 & 73.8/73.6 & 80.41/85.19 & 56.31 & 79.33  \\
    $1^{st}$-order Taylor & 76.31 & 28.88 & 83.26 & 74.64 & \textbf{{83.0/74.8}} & 76.7/76.6 & 80.09/85.29 & 57.76 & 81.20 \\
    Lookahead & 76.40 & 28.15 & 82.80 & 75.31 & 79.8/70.5 & 71.9/71.9 & 81.84/86.53 & 60.29 & 81.80 \\
    LAMP & 74.03 & 20.31 & 83.26 & 74.27 & 72.3/63.7 & 73.7/74.1 & 79.32/85.07 & 58.84 & 81.09 \\  
    \midrule
    \multicolumn{10}{l}{\textbf{Proposed Methodology}} \\
    \midrule
    $L_2$-MBP + SDP-COS  & 77.83 & 31.80 & 86.00 & 75.68 & 81.6/72.2 & 76.4/76.3 & 81.39/86.68 & 61.73 & 83.07  \\
    $L_2$-MBP + SDP-KLD & 78.34 & 36.74 & \textbf{{87.96}} & 77.94  & 80.5/68.2 & 77.1/77.3 & 83.21/85.58 & 63.18 & 83.54 \\
    $L_2$-MBP + SDP-CC & \textbf{{78.90}} & \textbf{{36.77}} & 87.84 & \textbf{{78.04}} & 81.1/71.0 & \textbf{77.3/77.5} & \textbf{{83.79}/86.37} & 62.64 & \textbf{{84.20}}  \\

    \bottomrule[1.25pt]
    \multicolumn{10}{l}{BERT- results reported from ~\citet{sanh2019distilbert,jiao2019tinybert,turc2019well} and MNLI results are for the matched dataset.}
        \end{tabular}
        }
        \caption{\textbf{GLUE benchmark results for pruned models @10\% (or @20\%) remaining weights.}}
        \label{tab:glue_prune}
    \end{center}
   \vspace{-0.6cm}
\end{table}

\iffalse
% STS-B @ 20\%
random 37.6/36.0
l0    79.1/79.0
l2    79.2/79.0
l2-glob 79.0/78.8
Grad    73.9/73.8
Taylor 76.8/76.6
LookA  77.1/77.3 
Lamp   71.5/72.1
----------------
COS-KD 79.3/79.1
KLD-KD 78.6/78.4
CC-KD  79.5/79.3
\fi

\begin{figure*}[ht]
\centering
\begin{subfigure}{.45\textwidth}
  \centering
  \includegraphics[width=1.\linewidth]{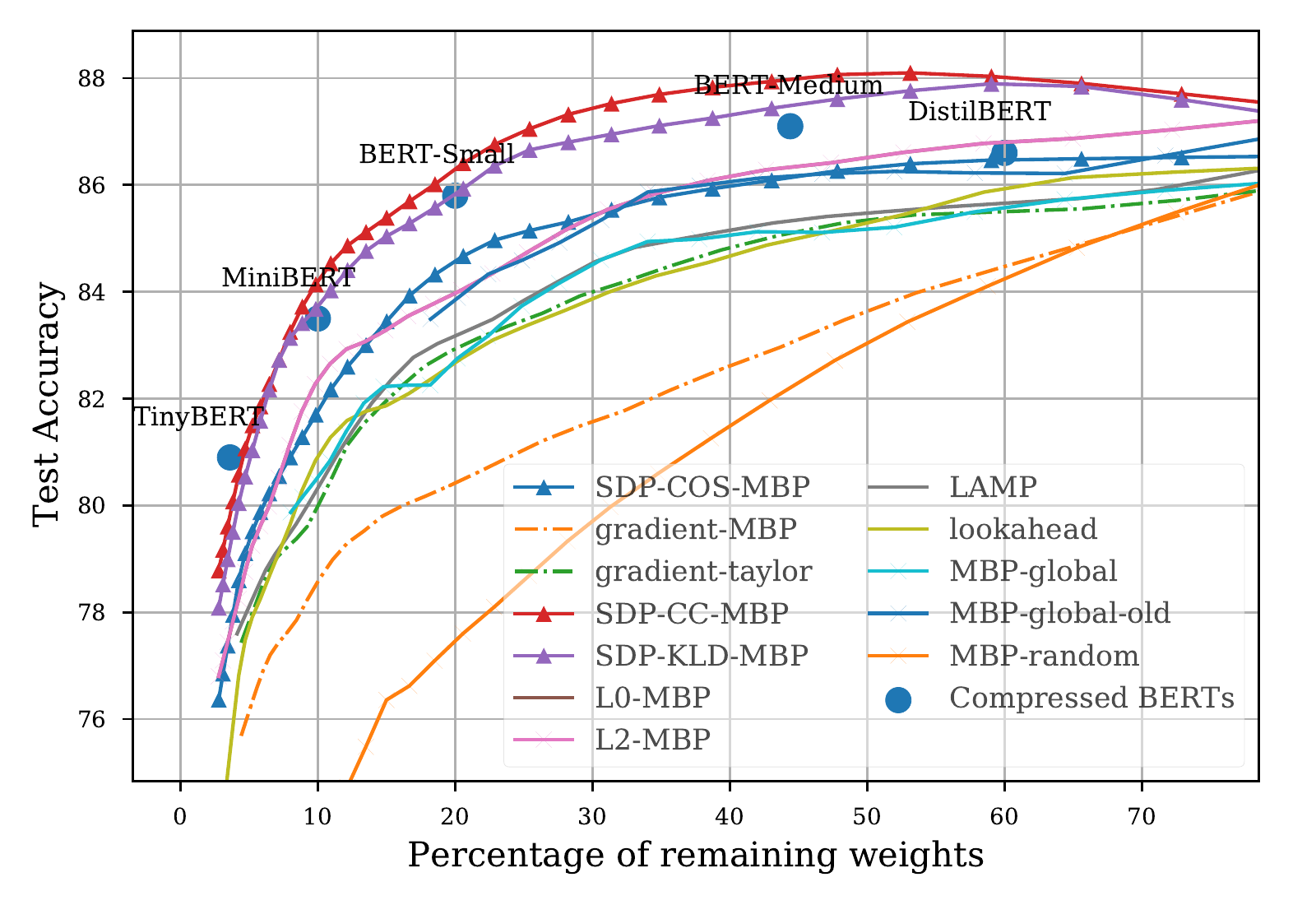}
  \caption{Question Answering NLI (QNLI)}
  \label{fig:qnli}
\end{subfigure}%
\begin{subfigure}{.45\textwidth}
  \centering
  \includegraphics[width=1.\linewidth]{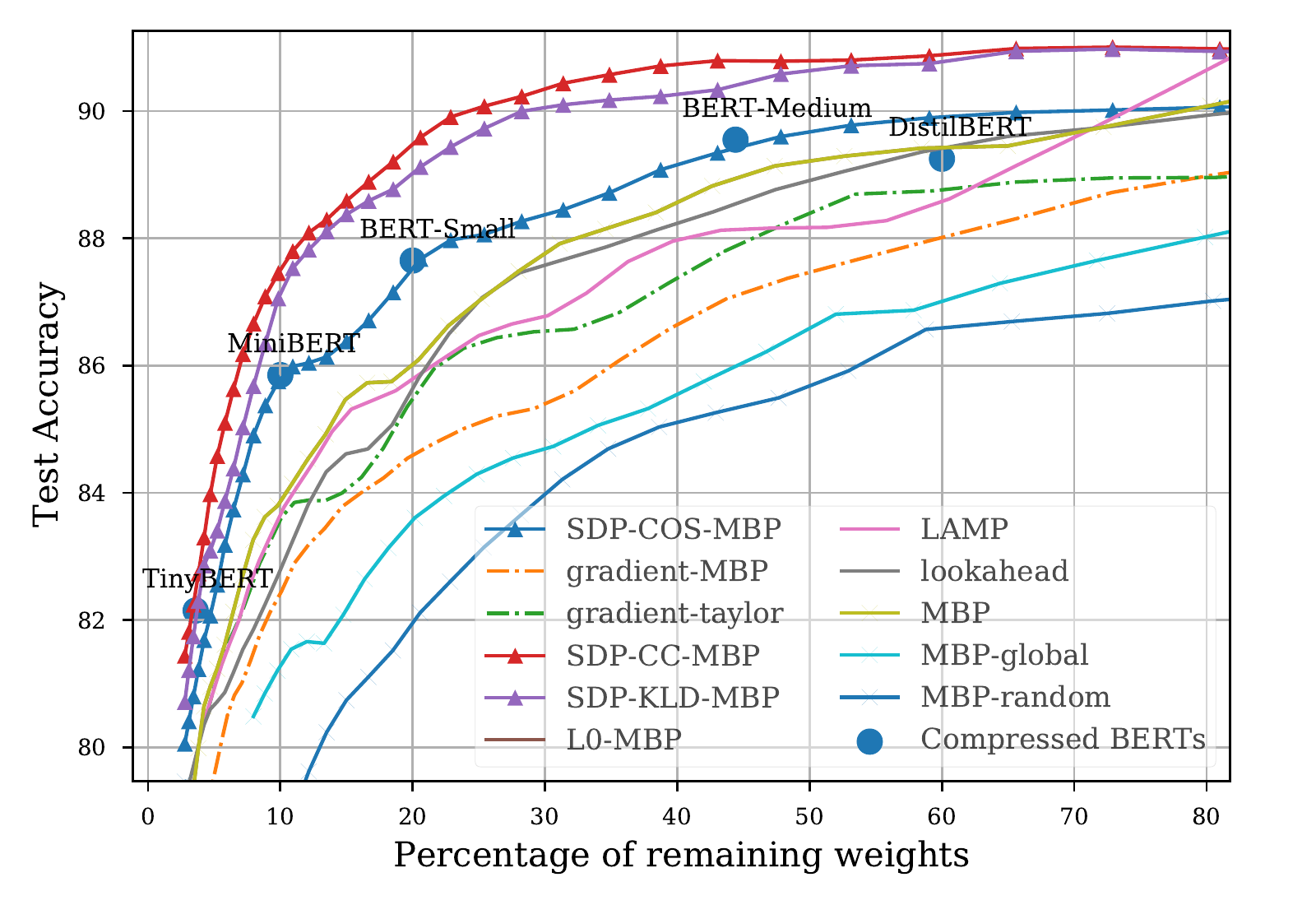}
  \caption{Sentiment Analysis (SST-2)}
  \label{fig:sst}
\end{subfigure}
\hfill
\begin{subfigure}{.45\textwidth}
  \centering
  \includegraphics[width=1.\linewidth]{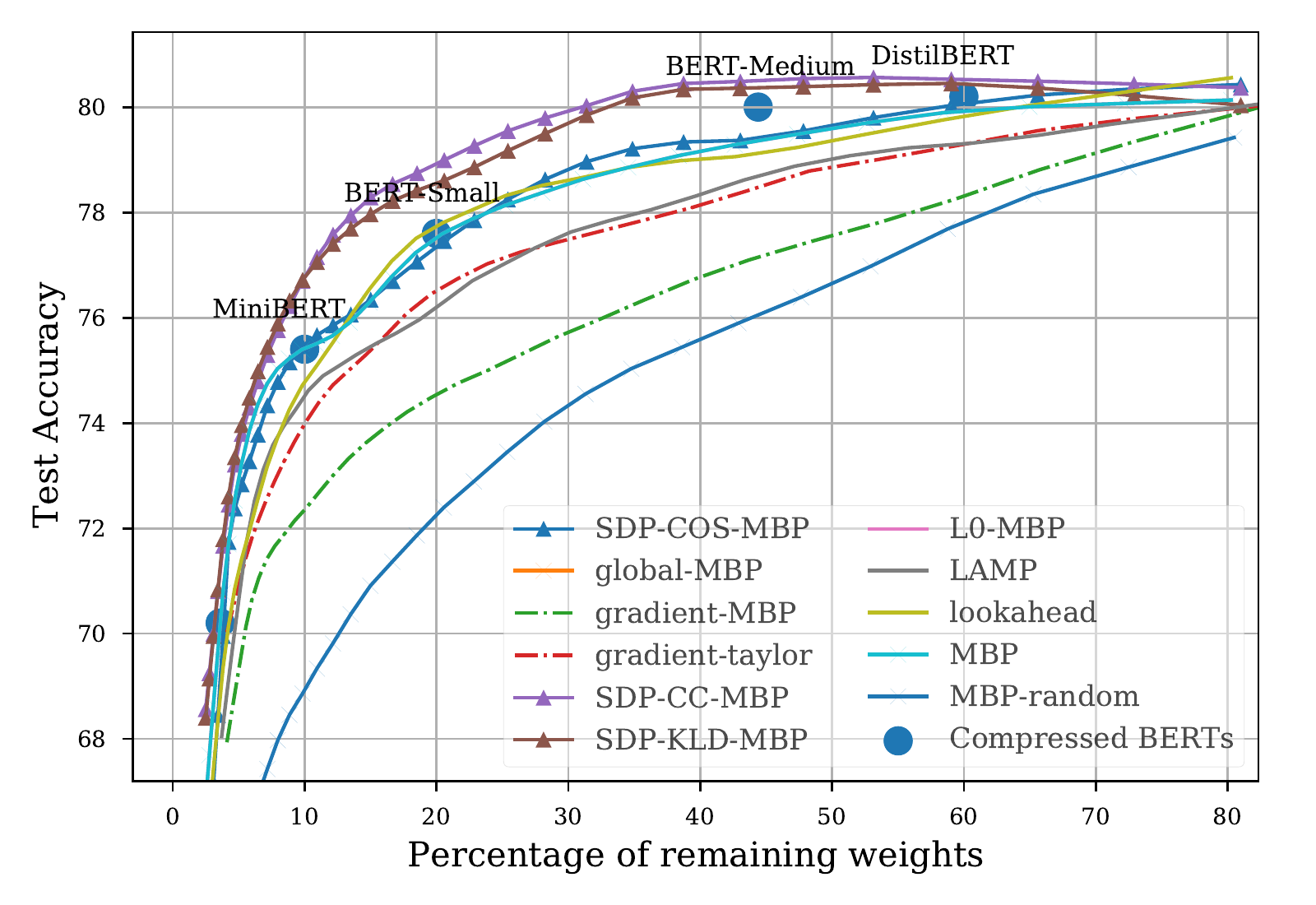}
  \caption{Multi-Genre NLI (MNLI)}
  \label{fig:mnli}
\end{subfigure}%
\begin{subfigure}{.45\textwidth}
  \centering
  \includegraphics[width=1.\linewidth]{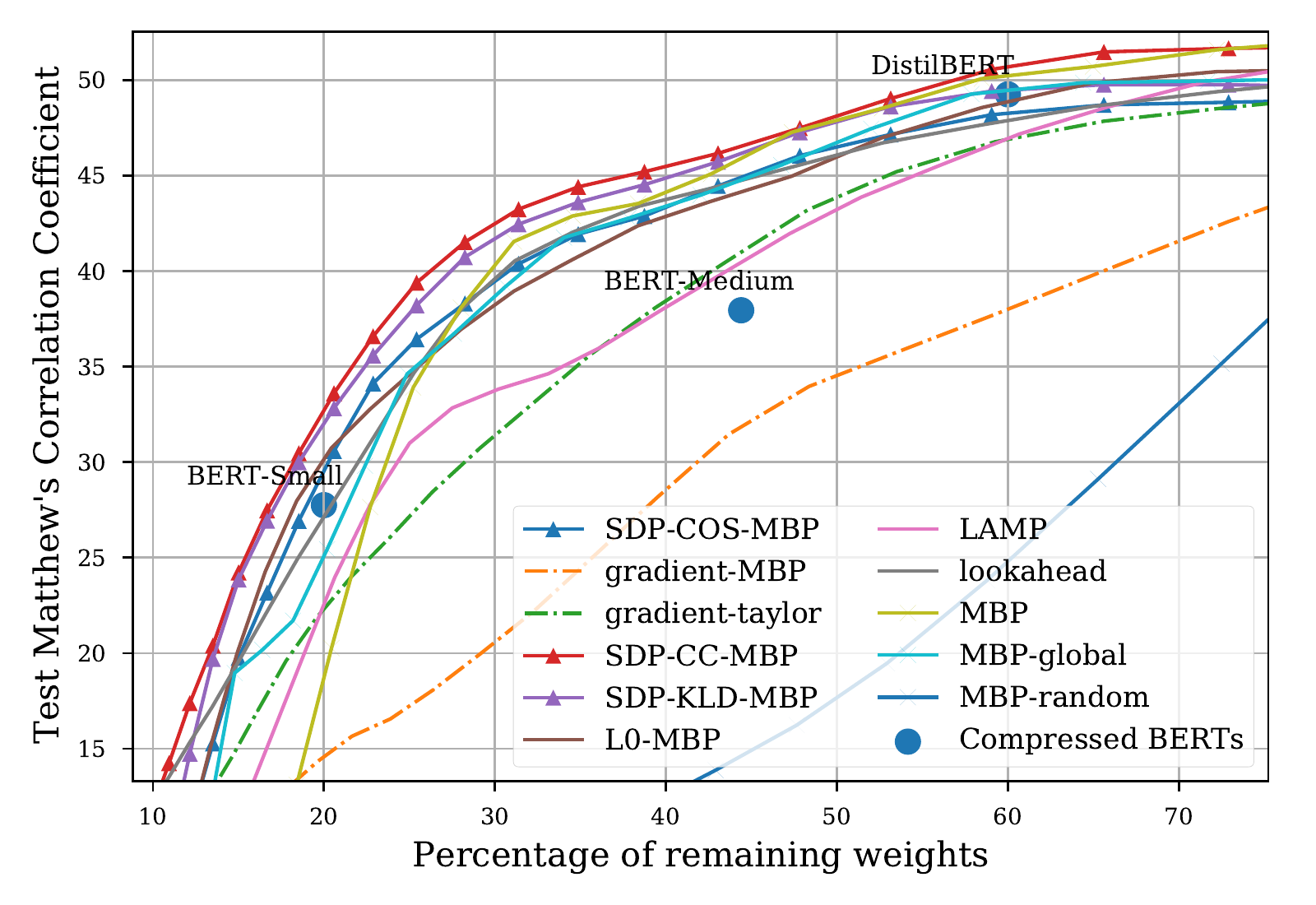}
  \caption{Linguistic Acceptability (CoLA)}
  \label{fig:cola}
\end{subfigure}
\caption{\textbf{Iterative Pruning Test Performance on GLUE tasks.}}
\label{fig:glue_pruning_methods}

\end{figure*}

Smaller dense versions of BERT require more labelled data in order to compete with unstructured MBP and higher-order pruning methods such as $1^{st}$ order Taylor series and Lookahead pruning. For example, we see BERT-Mini (@10\%) shows competitive test accuracy with our proposed SDP-CC on QNLI, MNLI and QQP, the three datasets with the most training samples (105k, 393k and 364k respectively). Overall, $L_2-$MBP + SDP-CC achieves the highest GLUE score for all models at 10\% remaining weights when compared to BERT-Base parameter count. Moreover, we find that $L_2$-MBP + SDP-CC achieves best performance for 5 of the 8 tasks, with 1 of the remaining 3 being from $L_2$MBP+SDP-KLD. This suggests that redundancy reduction via a cross-correlation objective is useful for SDP and clearly improve over SDP-COS which does not minimize correlations between off-diagonal terms.
\autoref{fig:glue_pruning_methods} shows the performance across all pruning steps. Interestingly, for QNLI we observe the performance notably improves between 30-70\% for SDP-CC and SDP-KLD. For SST-2, we observe a significant gap between SDP-KLD and SDP-CC compared to the pruning baselines and smaller versions of BERT, while TinyBERT becomes competitive at extreme compression rates ($<$4\%).
% Textual Similarity (STS-B)

\begin{figure*}[ht]
\begin{subfigure}{.33\textwidth}
  \centering
  \includegraphics[width=.975\linewidth]{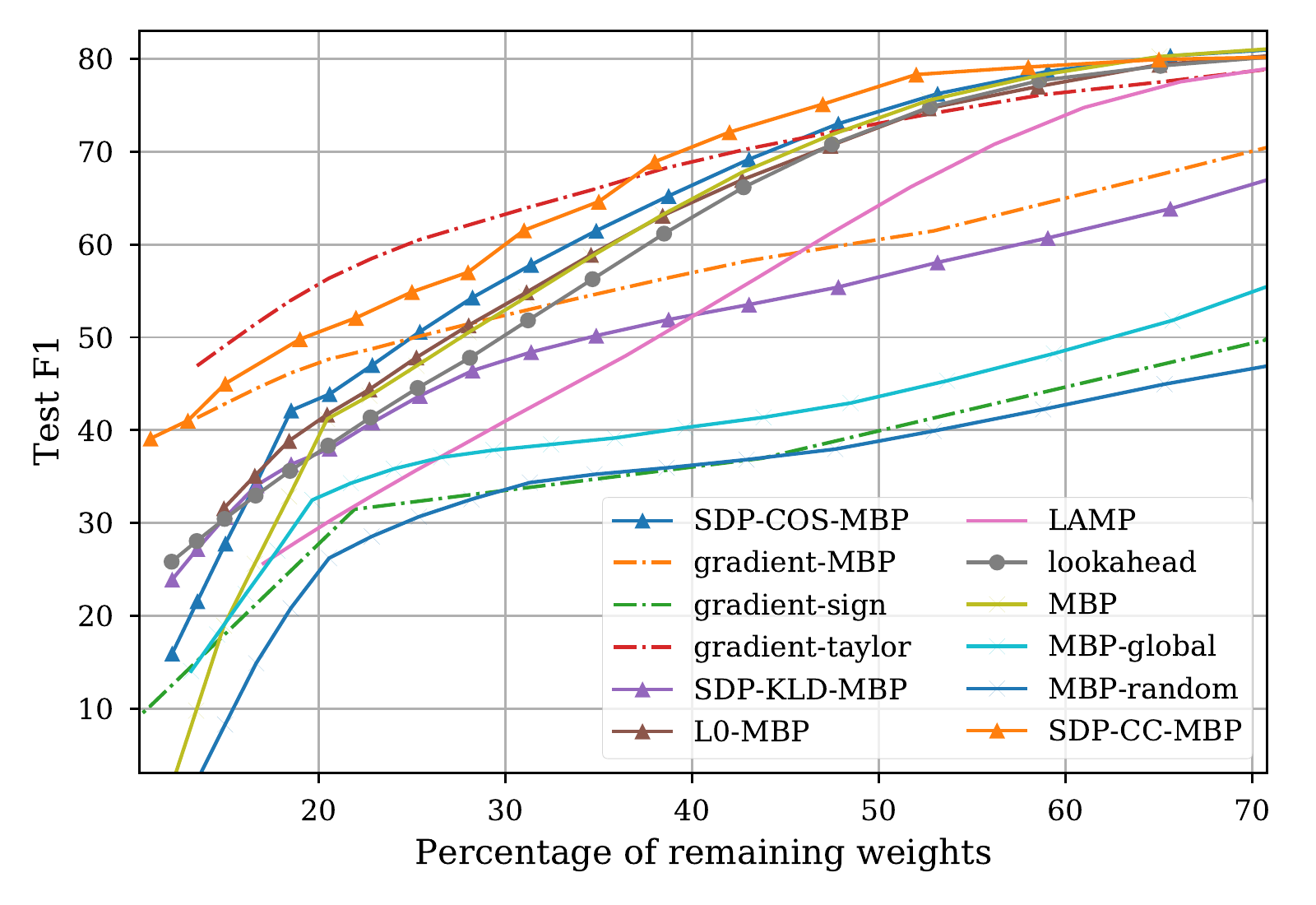}
  \caption{Named Entity Recognition}
  \label{fig:pos}
\end{subfigure}%
\begin{subfigure}{.33\textwidth}
  \centering
  \includegraphics[width=.975\linewidth]{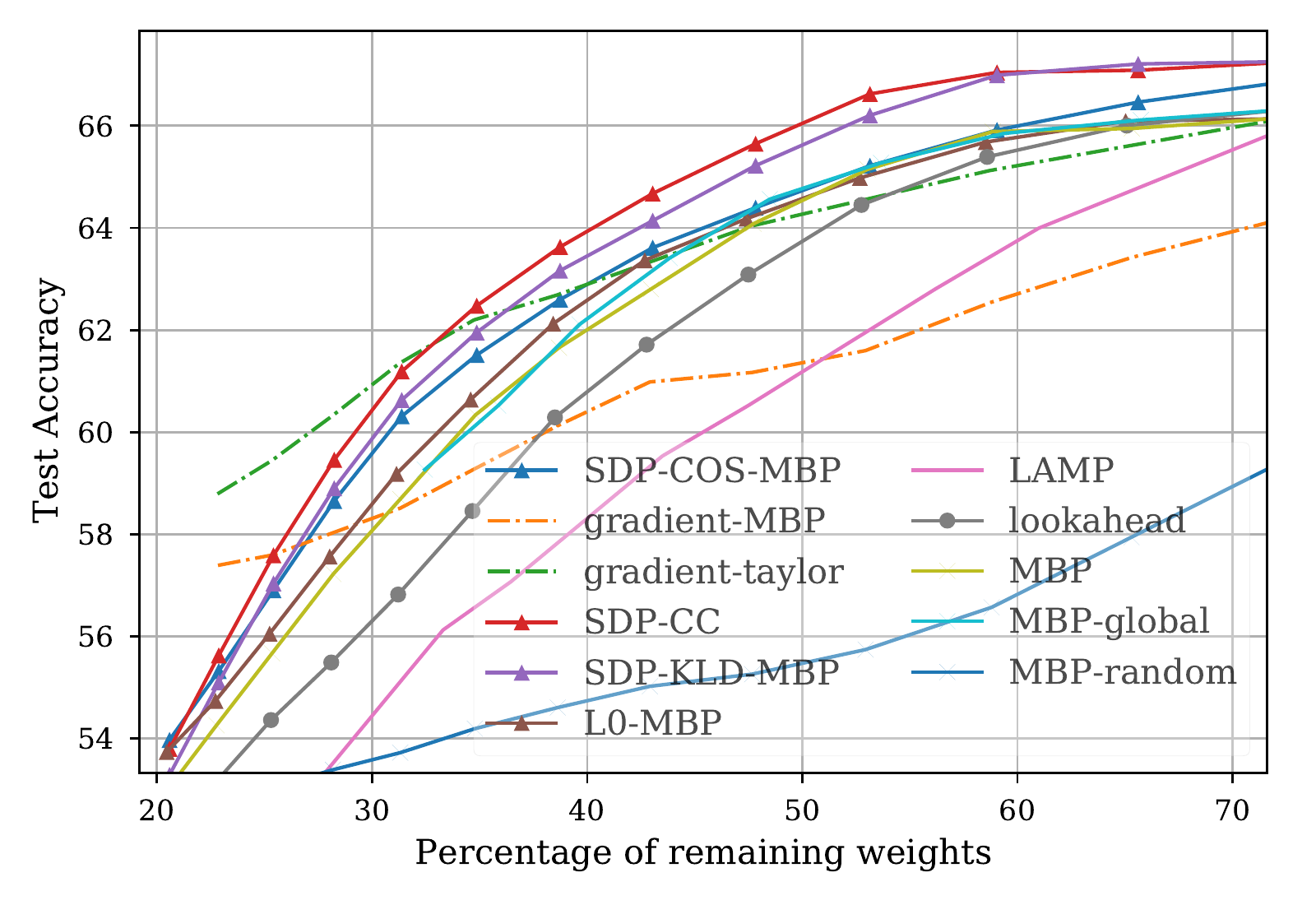}
  \caption{Query-Ad Matching}
  \label{fig:ner}
\end{subfigure}%
\begin{subfigure}{.33\textwidth}
  \centering
  \includegraphics[width=.975\linewidth]{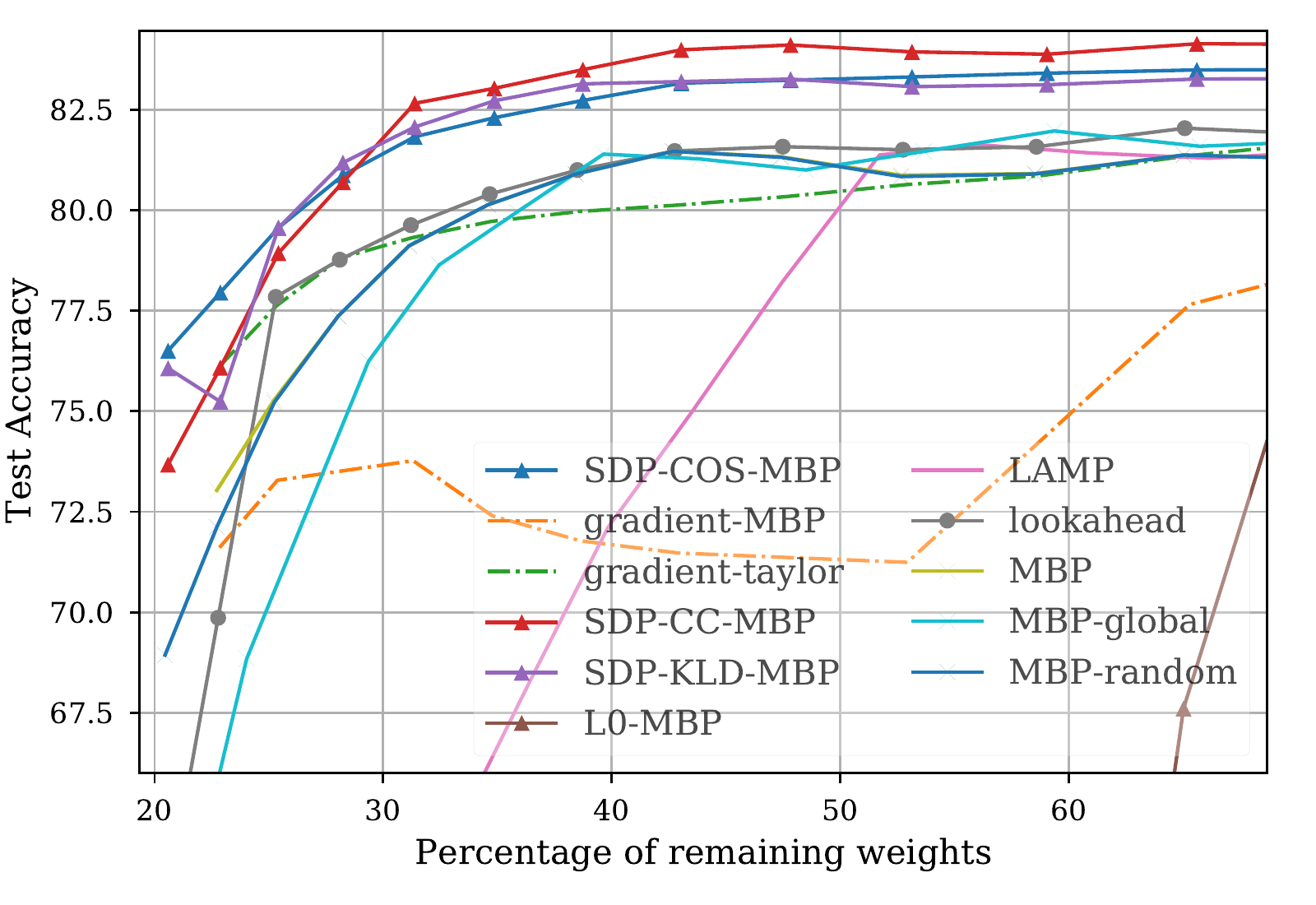}
  \caption{News Classification}
  \label{fig:news}
\end{subfigure}
\hfill
\begin{subfigure}{.33\textwidth}
  \centering
  \includegraphics[width=.975\linewidth]{images/xglue/seaborn/qam.pdf}
  \caption{Question Answer Matching}
  \label{fig:qam}
\end{subfigure}%
\begin{subfigure}{.33\textwidth}
  \centering
  \includegraphics[width=.975\linewidth]{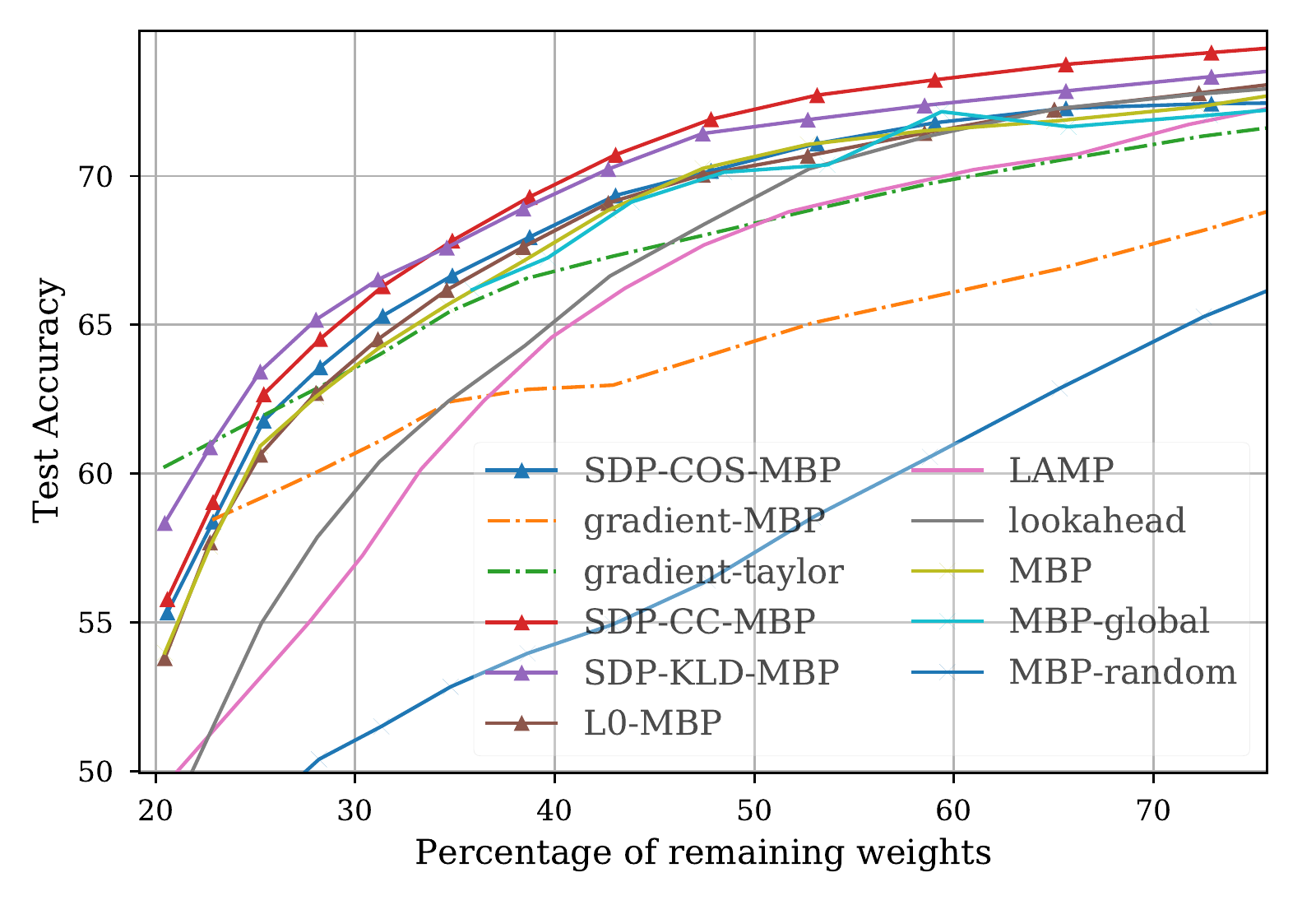}
  \caption{XNLI}
  \label{fig:xnli}
\end{subfigure}%
\begin{subfigure}{.33\textwidth}
  \centering
  \includegraphics[width=.975\linewidth]{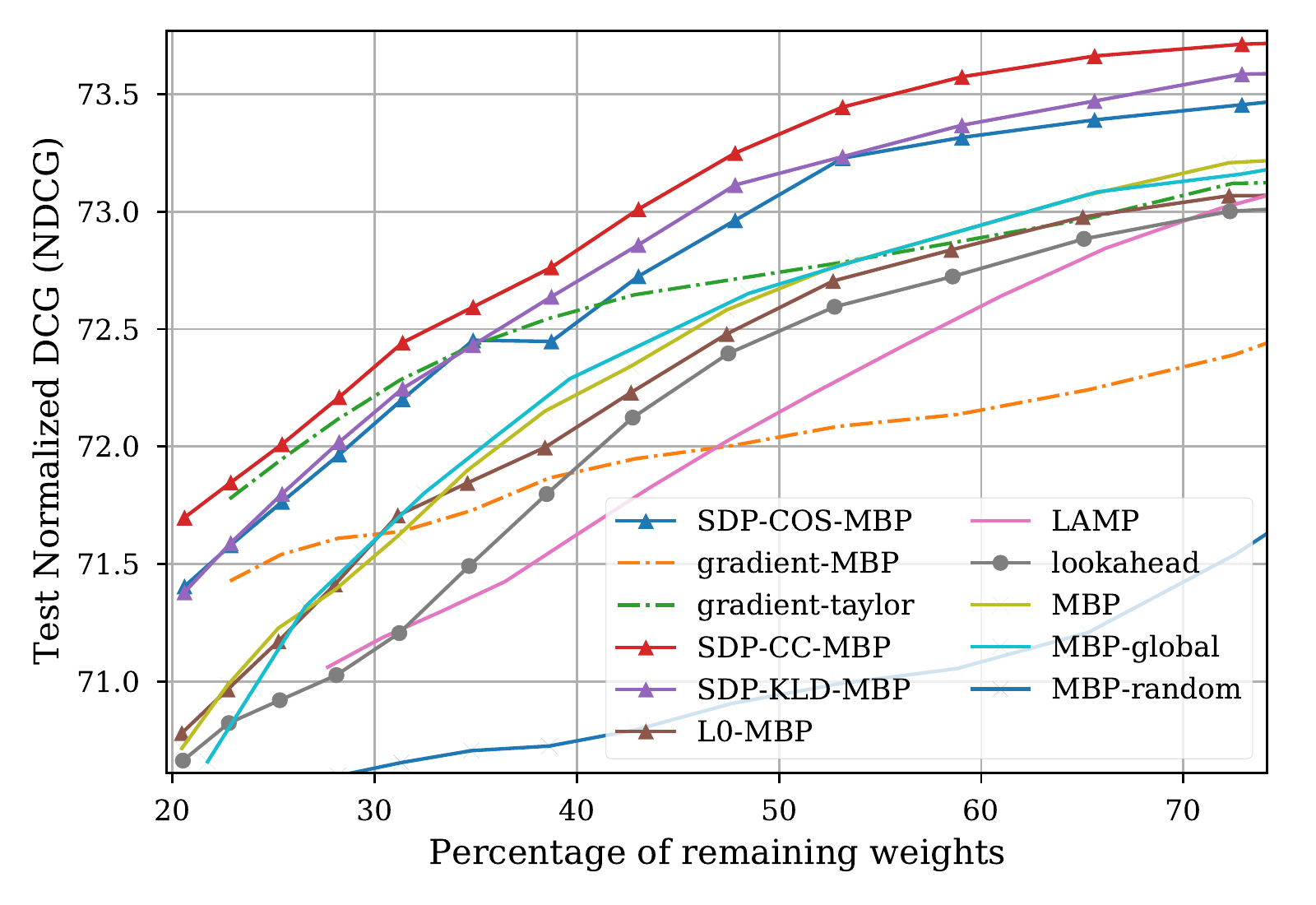}
  \caption{Web-Page Ranking}
  \label{fig:wpr}
\end{subfigure}
\vspace*{-2mm}
\caption{\textbf{Zero-Shot Results After Iteratively Fine-Pruning XLM-R$_{\mathrm{Base}}$ on XGLUE tasks}.}
\label{fig:xglue_pruning_methods}
\vspace*{-1mm}
\end{figure*}

%\paragraph{XGLUE Pruning Results.}
\textbf{Pruning Results on {\em XGLUE}.}
We show the per task test performance and the {\em average task understanding} score on XGLUE for pruning baselines and our proposed SDP approaches in Table \ref{tab:ft_all}. Our proposed cross-correlation objective for SDP again achieves the best average (Avg.) score and achieves the best task performance in 6 out of 8 tasks, while standard SDP-KLD achieves best performance on one (news classification) of the remaining two. Most notably, we outperform methods which use higher order gradient information ($1^{st}$-order Taylor) at 30\% remaining weights, which tends to be a point at which XLM-R$_{\mathrm{Base}}$ begins to degrade performance below 10\% of the original fine-tuned test performance for SDP methods and competitive baselines. In~\autoref{fig:xglue_pruning_methods}, we can observe this trend from the various tasks within XGLUE. We note that the number of training samples used for retraining plays an important role in the rate of performance degradation. For example, of the 6 presented XGLUE tasks, NER has the lowest number of training samples (15k) of all XGLUE tasks and also degrades the fastest in performance (from 90\% to 50\%  Test F1 at 30\% remaining weights). In comparison, XNLI has the most training samples for retraining (433k) and maintains performance relatively well, keeping within 10\% of the original fine-tuned model at 30\% remaining weights. 

% the number of zero-shot languages used to evaluate cross-lingual transfer and

\begin{table*}[t]
\begin{center}
    % \scriptsize
    \resizebox{1.0\linewidth}{!}{
    \begin{tabular}[b]{l|cccccccc|l}
    \toprule[1.25pt]
    \textbf{Prune Method} & \textbf{XNLI} & \textbf{NC} & \textbf{NER} & \textbf{PAWSX} & \textbf{POS} & \textbf{QAM} & \textbf{QADSM} & \textbf{WPR} &  \textbf{Avg.} \\ 
    \midrule
%NER	  POS   NC    MLQA  XNLI  PAWS-X   QADSM  WPR   QAM   AV
%79.7  79.6  83.5  66.0  75.3  90.1	   68.4	  73.9  68.9  76.1

% mlqa - 63.7 / 46.3
    XLM-R$_{\text{Base}}$ & 73.48 & 80.10 & 82.60 & 89.24 & 80.34 & 68.56 & 68.06 & 73.32 & 76.96 \\
    \midrule
    Random & 51.22 & 70.19 & 38.19 & 57.37 & 52.57 & 53.85 & 52.34 & 70.69 & 55.80  \\
    Global-Random & 50.97 & 69.88 & 38.30 & 56.74 & 53.02 & 54.02 & 53.49 & 69.11 & 55.69  \\
    $L_0$-MBP & 64.75 & 78.98 & 56.22 & 72.09 & 71.38 & 59.31 & 53.35 & 71.70 & 65.97 \\
    $L_2$-MBP & 64.30 & 78.79 & 54.43 & 77.99 & 70.68 & 59.24 & 60.33 & 71.52 & 67.16 \\
    $L_2$-Global-MBP & 65.12 & 78.64 & 54.47 & 79.13 & 71.37 & 59.26 & 60.61 & 71.80 & 67.55  \\
    $L_2$-Gradient-MBP & 61.11 & 73.77 & 53.25 & 79.56 & 65.89 & 57.35 & 59.33 & 71.59 & 65.23 \\
    $1^{st}$-order Taylor & 64.26 & 79.34 & 63.60 & \textbf{82.83} & 68.94 & 61.69 & 62.42 & 72.28 & 69.09 \\
    % LayerDrop & 57.20 & 73.20 & 52.38 & 71.01 & 56.73 & 57.31 & 58.03 & 69.10 & 61.87 \\
    Lookahead & 60.84 & 79.18 & 54.44 & 71.05 & 68.76 & 55.94 & 53.41 & 71.26 & 64.36 \\
    LAMP & 58.04 & 63.64 & 51.92 & 66.05 & 67.43 & 55.36 & 52.42 & 71.09 & 60.74 \\
    \midrule
     \midrule
    $L_2$-MBP + SDP-COS & 64.96 & 79.02 & 62.77 & 78.70 & 72.88 & 60.21 & 60.94 & 72.04 & 68.94 \\
    $L_2$-MBP + SDP-KLD & 65.94 & \textbf{80.72} & 64.50 & 79.25 & 73.18 & 61.66 & 61.09 & 71.84 & \textbf{69.77} \\
    $L_2$-MBP + SDP-CC & \textbf{66.47} & 79.73 & \textbf{66.34} & 80.03 & \textbf{73.45} & \textbf{63.73} & \textbf{62.78} & \textbf{72.59} & {\textbf{70.76}} \\
    % $L_2$-MBP + SDP-VID & 1 & 2 & 3 & 4 & 5 & 6 & 7 & 8 & 9 \\
    %\midrule
    %Global-MBP + SDP-VID & 1 & 2 & 3 & 4 & 5 & 6 & 7 & 8 & 9 \\
    %Global-MBP + SDP-COS & 1 & 2 & 3 & 4 & 5 & 6 & 7 & 8 & 9 \\
    %Global-MBP + SDP-KLD & 1 & 2 & 3 & 4 & 5 & 6 & 7 & 8 & 9 \\
    %Global-MBP + SDP-CC & 1 & 2 & 3 & 4 & 5 & 6 & 7 & 8 & 9 \\
    \bottomrule[1.25pt]
    \end{tabular}
    }
    \caption{\textbf{XGLUE Iterative Pruning @ 30\% Remaining Weights of XLM-R$_{\mathrm{base}}$} - Zero Shot Cross-Lingual Performance Per Task and Overall Average Score (Avg).}
    \label{tab:ft_all}
\end{center}
\vspace*{-6mm}
\end{table*}

\textbf{Summary of Results.} From our experiments on GLUE and XGLUE task, we find that SDP consistently outperforms pruning, KD and smaller BERT baselines. SDP-KLD and SDP-CC both outperform larger sized BERT models (BERT-Small), somewhat surprisingly, given that BERT-Small (and the remaining BERT models) have the advantage of large-scale self-supervised pretraining, while pruning only has supervision from the downstream task. For NER in XGLUE, higher order pruning methods such as Taylor-Series pruning have an advantage at high compression rates mainly due to lack of training samples (only 15k). Apart from this low training sample regime, SDP with MBP dominates at high compression rates both in standard and zero-shot settings.

\begin{figure*}[t!]
    \centering
    \begin{subfigure}[t]{0.5\textwidth}
        \centering
        \includegraphics[scale=0.453]{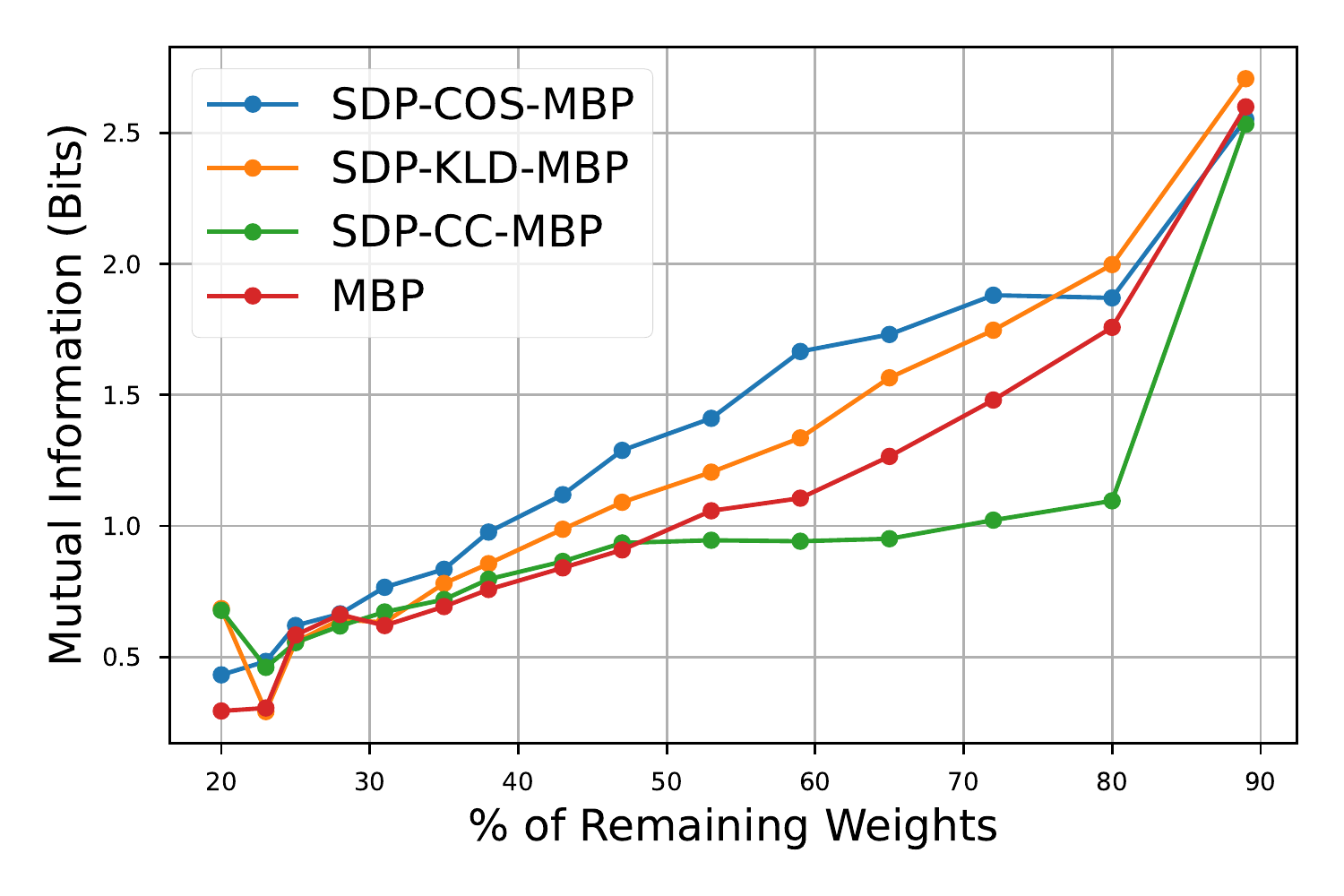}
        %\caption{Mutual Information Between [CLS] Hidden States}
    \end{subfigure}%
    ~ 
    \begin{subfigure}[t]{0.5\textwidth}
        \centering
        \includegraphics[scale=0.45]{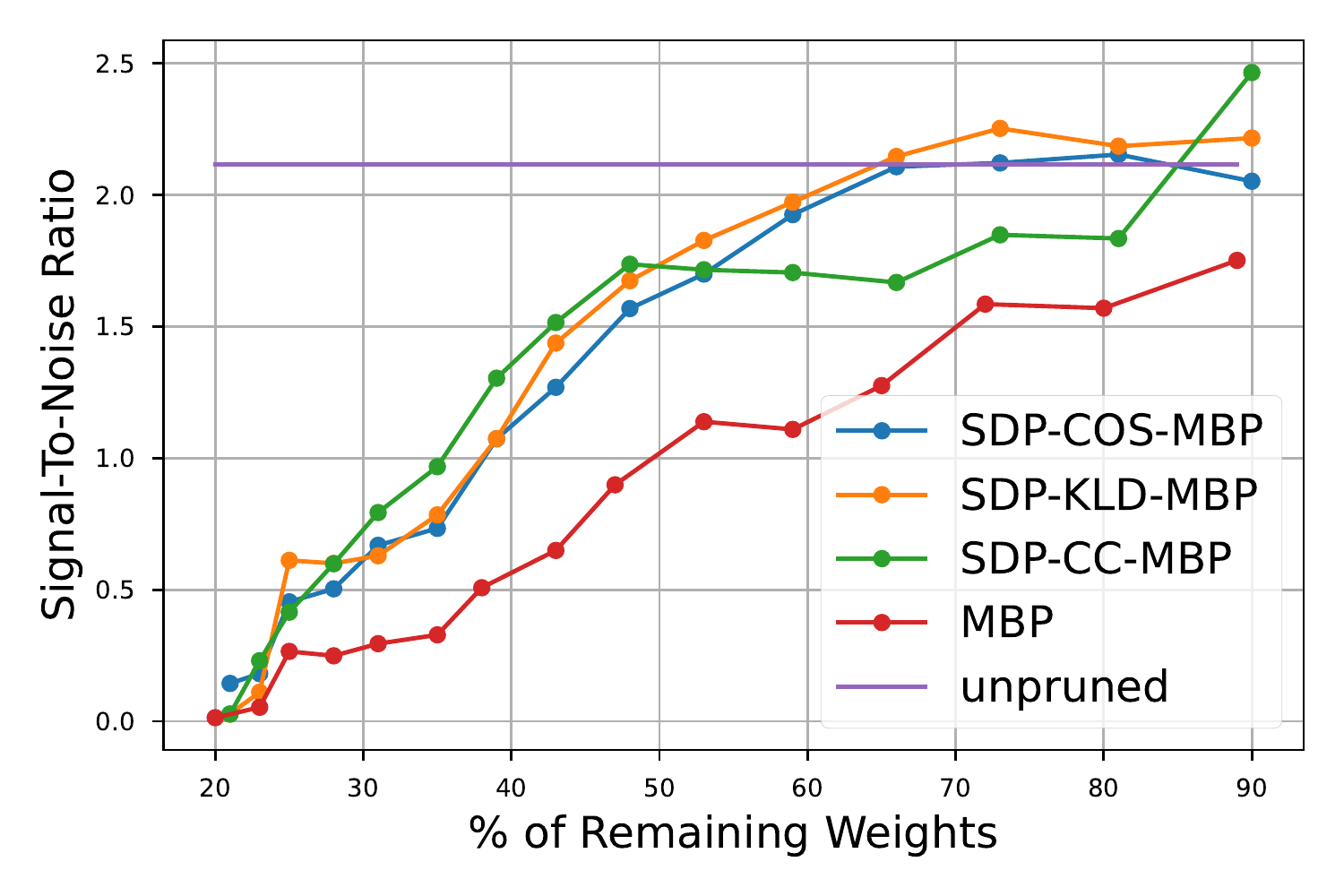}
        %\caption{Signal To Noise Ratio}
    \end{subfigure}
    \vspace*{-5mm}
    \caption{\textbf{Mutual Information Between Unpruned and Pruned Representations (left) and Signal-To-Noise Ratio (right) from PAWS-X Development Set Representations.}}\label{fig:mi_snr_pawsx}
    \vspace*{-2mm}
\end{figure*}

\textbf{Measuring Fidelity To The Fine-Tuned Model.}
We now analyse the empirical evidence that soft targets used in SDP may force higher fidelity with the representations of the fine-tuned model when compared to using MBP without self-distillation. As described in~\autoref{sec:sdp_improves_perf} we measure mutual dependencies between both representations of models with the best performing hyperparameter settings of $\alpha$, $\beta$ and the softmax temperature $\tau$. We note that increasing the temperature $\tau$ translates to ``peakier'' teacher logit distributions, encouraging SGD to learn a student with high fidelity to the teacher. From the LHS of~\autoref{fig:mi_snr_pawsx}, we can see that SDP models have higher mutual information (MI) with the teacher compared to MBP, which performs worse for PAWS-X (similar on remaining tasks, not shown for brevity). In fact, the rank order of the best performing pruned models at each pruning step has a direct correlation with MI, e.g., SDP-COS-MBP maintains highest MI and the highest test accuracy for PAWS-X for the same $\alpha$. However, too high fidelity ($\alpha=1.$) led to worse generalization compared to a balance between the task provided targets and the teacher logits ($\alpha=0.5$).

\textbf{Self-Distilled Pruning Increases Class Separability and The Signal-To-Noise Ratio (SNR).}
We also find that the SNR is increased at each pruning step as formulated in~\autoref{sec:snr}. From this observation, we find that {\em SDP-CC-MBP} using cross-correlation loss does particularly well in the 30\%-50\% remaining weights range. More generally, all 3 SDP losses clearly lead to better class separability and class compactness across all pruning steps compared to MBP (i.e., no self-distillation). 
% We also find that this separation is also reflected in the mutual information between the SDP representations and the original network representations when comparing the MI score between standard pruning representations and original network representations.

\iffalse
\begin{figure*}[ht]
    \centering
    \includegraphics[scale=0.5]{images/xglue/collab//pawsx.png}
    \caption{Class Separability t-SNE embeddings visualization of KLD-SDP (top left) and magnitude pruning (top right), Signal-To-Noise Ratio (bottom left) and Representation Correlation (bottom right)}
    \label{fig:collab_prune_pawsx}
\end{figure*}
\begin{figure}
    \centering
    \includegraphics[scale=0.425]{images/xglue/class_sep/pawsx_cc_kld_magnitude.png}
    \caption{Class Separability }
    \label{fig:seq_class_sep}
\end{figure}
\fi

\textbf{Self-Distilled Pruning Recovers Faster Performance Degradation Directly After Pruning Steps.} 
\begin{wrapfigure}[14]{r}{7cm}
    \vspace*{-4mm}
    \centering
    \includegraphics[scale=0.56]{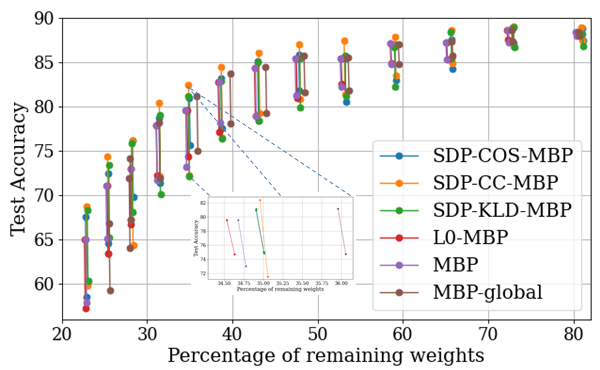}
    \vspace*{-5mm}
    \caption{\textbf{PAWS-X: Self-Distilled Pruning Leads to Better Performance Recovery.}}
    \label{fig:sdp_perf_recovery}
\end{wrapfigure}
\autoref{fig:sdp_perf_recovery} shows how SDP with Magnitude pruning (SDP-MBP) recovers during training in between pruning steps. The top of each vertical bar is the recovery development accuracy and the bottom is the initial performance degradation prior to retrainng. We see that SDP pruned models degrade in performance more than magnitude pruning without self-distillation. This suggests that SDP-MBP may force weights to be closer, as there is more initial performance degradation if weights are not driven to zero. However, the recovery is faster. This may be explained by recent work that suggests the stability generalization tradeoff~\citep{bartoldson2019generalization}.

\vspace{-0.6em}
\section{Conclusion}
\vspace{-0.5em}
In this paper, we proposed a novel {\em self-distillation} based pruning technique based on a {\em cross-correlation} objective. We extensively studied the confluence between pruning and self-distillation for masked language models and its enhanced utility on downstream tasks in both monolingual and multi-lingual settings. We find that self-distillation aids in recovering directly after pruning in iterative magnitude-based pruning, increases representational fidelity with the unpruned model and implicitly maximize the signal-to-noise ratio. Additionally, we find our cross-correlation based self-distillation pruning objective minimizes neuronal redundancy and achieves state-of-the-art in magnitude-based pruning baselines, and even 
%over existing baselines. SDP 
outperforms KD based smaller BERT models with more parameters.
 
%\subsubsection*{Acknowledgments}
%Use unnumbered third level headings for the acknowledgments. All acknowledgments, including those to funding agencies, go at the end of the paper.

\bibliography{iclr2022_conference}
\bibliographystyle{iclr2022_conference}

%\appendix
%\section{Appendix}
%You may include other additional sections here.

\end{document}

\maketitle

\section{Background}
Since our proposed method uses self-distillation to improve pruning performance, we briefly describe pruning and knowledge distillation.

\paragraph{Pruning}
Pruning weights is used to reduce the number of parameters in a pretrained DNN with the aim of reducing storage, model runtime and maintaining the same, or close to the performance original unpruned network. Iterative pruning prunes while retraining until the desired network size and accuracy tradeoff is met. Pruning is not carried out at random, but selected so that unimportant information about past experiences is discarded. 

\paragraph{Knowledge Distillation}
Knowledge distillation transfers the logits~\cite{hinton2015distilling} (or hidden representations~\cite{romero2014fitnets}) of a learned teacher network to a student network that and uses these logits as an additional soft target to optimize for. This typically leads to improved student network test performance, where the student network is relatively much smaller than the teacher network. Since the inception of knowledge distillation~\cite{bucilua2006model}, there has been various extensions. This includes distilling intermediate representations~\cite{romero2014fitnets}, distributions~\cite{huang2017like}, maximizing mutual information between student and teacher representations~\cite{ahn2019variational}, using pairwise interactions for improved knowledge distillation~\cite{park2019relational}, adversarial distillation~\cite{lan2018knowledge} and contrastive representation distillation~\cite{tian2019contrastive,neill2021semantically}.

\paragraph{Self-Distillation}
Self-distillation is a special case of knowledge distillation whereby the student network is the same size as the teacher network. Self-distillation has found to have a strong regularization effect as the student often generalizes better than the teacher even though they have the same capacity\cite{furlanello2018born,yang2019training,ahn2019variational}. The mechanisms by which self-distillation leads to improved generalization remains an open question. 

Recent work~\cite{stanton2021does} has shown that the discrepancy between the student and teacher outputs is mainly due to optimization difficulty, even in the self-distillation case where student and teacher have equal capacity and this discrepancy is heavily data-dependent. 

Allen et al.~\cite{allen2020towards} show that self-distillation can be viewed as implicitly combining ensemble and knowledge distillation to explain the improvement in test accuracy when dealing with multi-view data. The core idea is that the self-distillation objective results in the network learning a unique set of features through gradient descent that are distinct for the set of features learned from the originally trained model. Hence, these completementary features are implicitly learning distinct features similar to combining the outputs of independent models in an ensemble. 

% ~\citeauthor{gotmare2018closer} have empirically observed that the dark knowledge transferred by the teacher is localized mainly in higher layers and does not affect early (feature extraction) layers much.

% ~\citeauthor{furlanello2018born} interprets dark knowledge as importance weighting.~\citeauthor{dong2019distillation} shows that early-stopping is crucial for reaching dark-knowledge of self-distillation.~\citeauthor{abnar2020transferring} empirically study how inductive biases are transferred through distillation. Ideas similar to self-distillation have been used in areas besides modern machine learning but with different names such as diffusion and boosting in both the statistics and image processing communities~\cite{milanfar2012tour}.

In the proceeding sections we shed light on potential reasons why self-distilled models generalize better and how these insights pertain to improving pruned models.

\section{Related Research}
\paragraph{Regularization-based pruning}. The first group of relevant works is those applying regularization to learn sparsity. The most famous probably is to use $L_0$ or $L_1$ regularization~\cite{louizos2017learning,liu2017learning,ye2018rethinking} due to their sparsity-inducing nature. In addition, the common L2 regularization is also explored for approximated sparsity ~\cite{han2015learning,han2015compressing}. The early
papers focus more on unstructured pruning, which is beneficial to model compression yet not to acceleration. For structured pruning in favor of acceleration, Group-wise Brain Damage~\cite{lebedev2016fast} and SSL~\cite{wen2016learning} propose to use Group LASSO~\cite{yuan2006model} to learn regular sparsity, where the penalty strength is still kept in small scale because the penalty is uniformly applied to all the weights. To resolve this,~\citeauthor{ding2018auto} and~\cite{wang2019eigendamage} propose to employ different penalty factors for different weights, enabling large regularization. 

\paragraph{Importance-based pruning}. Importance-based pruning tries to establish certain advanced importance criteria that can reflect the true relative importance among weights as faithfully as possible. The pruned weights are usually decided immediately by some proposed formula instead of by training (although the whole pruning process can involve training, e.g., iterative pruning). The most widely used criterion is the magnitude-based: weight absolute value for unstructured pruning~\cite{han2015learning,han2015compressing} or L1/L2-norm for structured pruning~\cite{liu2017learning}. This heuristic criterion was proposed a long time ago~\cite{reed1993pruning} and has been argued to be inaccurate. In this respect, improvement mainly comes from using Hessian information to obtain a more accurate approximation of the increased loss when a weight is removed~\cite{lecun1990optimal,hassibi1993second}. Hessian is intractable to compute for large networks, so some methods (e.g.,~\citeauthor{wang2019eigendamage},~\citeauthor{singh2020woodfisher}) employ cheap approximation (such as K-FAC~\cite{martens2015optimizing}) to make the 2nd-order criteria tractable on deep networks. There is no a hard boundary between the importance-based and regularization-based. The difference mainly lies in their emphasis: Regularization-based method focuses more on an advanced penalty scheme so that the subsequent pruning criterion can be simple; while the importance-based one focus more on an advanced importance criterion itself. Meanwhile, regularization paradigm always involves iterative training, while the importance-based can be one-shot~\cite{lecun1990optimal,hassibi1993second,wang2019eigendamage} (no training for picking weights to prune) or involve iterative training~\cite{molchanov2016pruning,molchanov2019importance,ding2019global}.

\section{Details on Baselines}
\subsection{Regularization-Based Pruning Methods}
\paragraph{$L_0$ regularized pruning} - Since $\ell_0$ regularization of the weights is non-differentiable and thus cannot be used as a regularization term in the loss,~\citet{louizos2017learning} proposed to approximate which weights to set to 0 using non-negative stochastic gates (i.e hard concrete distribution), using
a differentiable binary concrete distribution and then transforming its sample with a hard sigmoid. The
parameters of the distribution over the gates can then be jointly optimized with the original network parameters.

Finally we note that movement pruning (and its soft variant) yield a similar update as $L_0$ regularization based pruning, another movement based pruning approach~\cite{louizos2017learning}. Instead of straight-through, L0 uses the hard-concrete distribution, where the mask $M$ is sampled for all $i, j$ with hyperparameters $b > 0, l < 0$, and $r > 1$:

\begin{equation}
\begin{gathered}
u \sim U(0, 1) \quad \bar{S}_{i,j} = \sigma(\log(u) - \log(1 - u) + S_{i,j} )/b \\
Z_{i,j} = (r - l)\bar{S}_{i,j} + l \quad M_{i,j} = \min(1, \mathrm{ReLU}(Z_{i,j}))    
\end{gathered}
\end{equation}

The expected $L_0$ norm has a closed form involving the parameters of the hard-concrete: $
\mathbb{E}(L_0) = \sum_{i,j} \sigma (\log S_{i,j} - b \log(-l/r)) $.

Thus, the weights and scores of the model can be optimized in an end-to-end fashion to minimize the sum of the training loss L and the expected $L_0$ penalty. A
coefficient $\lambda_{L_0}$ controls the $L_0$ penalty and indirectly the sparsity level. Gradients take a similar form:

\begin{equation}
\frac{\partial L}{\partial S_{i,j}} = \frac{\partial L}{\partial a_i} W_{i,j} x_j f(\bar{S}_{i,j} ) \quad \text{where} \quad f(S_{i,j} ) = \frac{r-l}{b} \bar{S}_{i,j} (1 - \bar{S}_{i,j} )\mathbf{1}{0 \leq Z_{i,j} \leq 1}    
\end{equation}

At test time, a non-stochastic estimation of the mask is used: $\tilde{M}=\min\Big(1,\mathrm{ReLU}\big((r-l)\sigma(S)+l\big)\Big)$ and weights multiplied by $0$ can simply be discarded.

\paragraph{Pruned Weight Overlap}
To analyse the main differences between SDP and standard iterative pruning, we analyse the overlap between which weights in each layer are chosen for pruning.

\paragraph{Weight Norms}
One well-established complexity measure of generalization is the Frobenius norm~\citep{neyshabur2017exploring} of the difference between representations from initialization to fine-tuned state. In the case of compression, we instead measure the distance between the fine-tuned representations to representations retrieved from a pruned network. We focus on the penultimate layer corresponding to the input CLS token in XLM-R$_{\mathrm{Base}}$.

\subsection{Teacher Labels Provide Higher Order Information For Magnitude Pruning}

\subsection{On Class Separability}
~\cite{muller2019does} have found that teachers trained with label smoothing act as worse teachers for student network than teachers learned without label smoothing in the knowledge distillation setting. We hypothesize that this is because the student benefits from the implicit class similarities given by the teacher logits which may become too flat when trained with label smoothing. In turn, this effects the transfer of knowledge w.r.t to class separability and convergence. 

\subsection{Generalization of Self-Distillation}

To begin, we define the Kullbeck-Leibler Divergence between the student and teacher distributions as the following:

\begin{gather}
    D_{\mathrm{KL}}(\vec{y}^T||\vec{y})=\sum_d \vec{y}^T_d\log\Big(\frac{\vec{y}^T_d}{\vec{y}_d}\Big)=\sum_d \vec{y}^T_d(\log \vec{y}^T_d - \log \vec{y}_d) = \\
    \sum_d \underbrace{\vec{y}^T_d\log \vec{y}^T_d}_{\text{entropic regularizer}} - \sum_d \underbrace{\vec{y}^T_d\log \vec{y}_d}_{\text{KD cross-entropy}}
\end{gather}

As the entropic regularizer doesn't include terms w.r.t $\theta$, we need only focus on the gradient of the KD cross-entropy loss which can be expressed as the \autoref{eq:ce}. 

\begin{equation}\label{eq:ce}
\ell_{ce} = - \log(q_y)= -\log\big(\frac{e^{z_y}}{\sum_j e^{z_j}}\big) = - z_y + \log \sum_j e^{z_j}    
\end{equation}

Calculating the derivative for each $z_i$:
\begin{gather}
\nabla_{zi}\ell_{\text{ce}} = \nabla_{z_i}(-z_y + \log \sum_j e^{z_j}) = 
\nabla_{z_i}\log \sum_j e^{z_j} - \nabla_{z_i}z_y = \\ \frac{1}{\sum_j e^{z_j}}\nabla_{z_i}\sum_j e^{z_j} - \nabla_{z_i}z_y = e^{z_i}\sum_j e^{z_j} - \nabla_{z_i}z_y = q_i - \nabla_{z_i}z_y= \\ 
q_i  - \mathbf{1}(y=i) \quad \text{where} (y=i) = 1 \quad \text{if} \quad y=i0 \quad \text{otherwise}    
\end{gather}

\bibliography{iclr2022_conference}
\bibliographystyle{iclr2022_conference}

\appendix
\section{Appendix}
You may include other additional sections here.